\documentclass{article}

\usepackage{adjustbox}
\usepackage{microtype}
\usepackage{graphicx}
\usepackage{subcaption}
\usepackage{booktabs} 

\usepackage{hyperref}

\usepackage[preprint]{icml2026}

\usepackage{amsmath}
\usepackage{amssymb}
\usepackage{mathtools}
\usepackage{amsthm}

\usepackage[capitalize,noabbrev]{cleveref}

\theoremstyle{plain}

\theoremstyle{definition}

\theoremstyle{remark}

\usepackage[textsize=tiny]{todonotes}

\icmltitlerunning{GAP3D: Generative Alignment for 3D Generation}

\begin{document}

\twocolumn[
  \icmltitle{GAP3D: Generative Alignment of VLM Latents \\ to Patch-Level Embeddings for 3D Generation}



  \icmlsetsymbol{equal}{$\dagger$}

  \begin{icmlauthorlist}
    \icmlauthor{Polytimi Anna Gkotsi}{yyy}
    \icmlauthor{Andrii Zadaianchuk}{equal,yyy}
    \icmlauthor{Mohammad Mahdi Derakhshani}{equal,yyy}
  \end{icmlauthorlist}

  \icmlaffiliation{yyy}{University of Amsterdam, The Netherlands}

  \icmlcorrespondingauthor{Polytimi Anna Gkotsi}{polytimi.gkotsi@student.uva.nl}

  \icmlkeywords{Vision-language models, 3D asset generation, Diffusion models, Representation alignment, Foundation model integration, Flow matching}

  \vskip 0.3in
]



\printAffiliationsAndNotice{\icmlEqualContribution}

\begin{abstract}
Recent approaches integrating vision-language models (VLMs) as prompt encoders for generative model conditioning typically rely on expensive end-to-end training or map features to compressed representations, discarding the dense spatial structure required for geometry-aware tasks like 3D asset generation. To address this, we propose GAP3D, a modular, diffusion-based approach that aligns VLM-generated latents directly to the complete, patch-level feature space of a pre-trained image encoder, enabling a frozen downstream generative model to utilize a VLM as prompt encoder while maintaining a spatially structured conditioning signal. Evaluated on 3D asset generation, our method bypasses the need for large-scale 3D data by training mainly on general-domain image-text pairs. It also exhibits emergent zero-shot capabilities for multimodal prompts, despite being trained exclusively on text input. Finally, while currently prioritizing high-level semantics over fine-grained detail, GAP3D demonstrates that the representation gap between VLM and image-encoder feature spaces can be partially bridged through diffusion-based alignment, taking the first steps towards a modular integration of foundation models through generative alignment to dense embedding spaces.\looseness-1

\end{abstract}
\section{Introduction}

Generative models require rich conditioning for precise control, yet standard text encoders like CLIP \cite{radford2021learningtransferablevisualmodels} and T5 \cite{10.5555/3455716.3455856} can struggle with overlong prompts, fine-grained attribute binding, or complex compositional structure \cite{han2025progressive,10847875, lee2023holisticevaluationtexttoimagemodels,wu2025qwenimagetechnicalreport}, constrained by their input length limitations, reliance on global semantics, or lack of spatial understanding, following their contrastive or text-only training objectives \cite{ li2024unimogunifiedimagegeneration,tan2024empiricalstudyanalysistexttoimage}. 

 In contrast, modern VLMs possess large representation capacity and strong multimodal reasoning capabilities. However, integrating them into generative pipelines remains challenging: Tightly coupled frameworks \cite{li2024unimogunifiedimagegeneration, li2025unifusionvisionlanguagemodelunified, wu2025qwenimagetechnicalreport,yang2025focusunifiedvisionlanguagemodeling}, require computationally expensive re-training or the development of entirely new generative backbones, lacking compatibility with existing pre-trained systems and requiring large amounts of domain-specific data, whereas modular approaches that align VLM features to the input spaces of frozen, pre-trained diffusion models \cite{pan2024kosmosggeneratingimagescontext, dong2024dreamllmsynergisticmultimodalcomprehension, chen2025blip3ofamilyfullyopen,  koh2023generatingimagesmultimodallanguage,mi2025i} primarily target pooled image features or text-token embeddings, discarding the patch-level structure needed for fine-grained appearance and geometry. 

This raises a fundamental question: can VLM representations be mapped not to compressed semantic vectors, but to dense, patch-level conditioning spaces already utilized by frozen generative models? Such alignment could provide frozen generative models with conditioning signals that are both semantically grounded (through VLM reasoning) and spatially detailed (through patch-level structure), directly benefiting tasks that require precise geometric control (e.g. defining the shape of a 3D object's handle). 

We investigate this through generative latent feature mapping, using a diffusion-based transformer to map VLM-generated latents to the complete embedding space of a pre-trained image encoder, including classification tokens, register tokens, and all spatial patch embeddings. We select a generative approach over deterministic regression, because mapping abstract semantic features such as those obtained  by a VLM to dense visual representations requires synthesizing geometry and appearance details that may not be explicitly present in the input prompt, rendering the alignment task non-deterministic.  We call our method GAP3D\footnote{The accompanying code is released at \url{https://github.com/PolyannaG/GAP3D}.} (\textbf{G}enerative \textbf{A}lignment of VLM latents to \textbf{P}atch-level embeddings for \textbf{3D} generation).\looseness-1

We evaluate our approach using embedding cosine similarity and zero-shot text-to-image retrieval, but focus primarily on 3D asset generation. As 3D tasks require precise geometry and attribute binding, features often lost in compressed embeddings, this domain is ideal for verifying dense alignment. Furthermore, the scarcity of 3D data and the training cost of 3D generative models makes our modular approach especially valuable. For this purpose, we append trainable embeddings to a frozen VLM's input, producing a sequence of latent representations. These condition a diffusion transformer, trained via flow matching to generate embeddings in the representation space of the DINOv2 \cite{oquab2024dinov2learningrobustvisual} image encoder. The generated embeddings then serve as conditioning for an existing frozen image-to-3D generative model (TRELLIS \cite{xiang2024structured}), originally trained with the specific encoder. GAP3D builds upon the generative alignment paradigm introduced by BLIP3-o \cite{chen2025blip3ofamilyfullyopen}, but extends it from a small set of compressed, pooled tokens to the complete, high-dimensional, and spatially structured patch space of a purely visual encoder.

Our experiments demonstrate alignment of VLM-generated latents with the target image encoder's patch-level embeddings, showing that independently pre-trained foundation models can be connected in high-dimensional spaces through diffusion. Our main contributions are:

\begin{itemize}
    \item \textbf{Dense generative representation alignment}: We show that VLM-generated features can be aligned to patch-level embedding spaces of pre-trained image encoders through flow-matching, moving beyond compressed or semantic representations. We thus establish the feasibility of connecting pre-trained foundation models through generative representation alignment even in high-dimensional feature spaces.

    \item \textbf{Modular 3D asset generation}: We apply our method to 3D asset generation, extending prior work on VLM-based conditioning for generation beyond the image domain, and enabling modular text-to-3D generation and emergent multimodal conditioning, without end-to-end retraining. 
    \item \textbf{Domain adaptation insights}: We show that domain-specific fine-tuning is critical for bridging the distribution shift between natural images and synthetic 3D renderings, reducing artifacts and ensuring high-fidelity outputs in such a modular pipeline.

\end{itemize}

\section{Related Work}

\subsection{Vision-Language Models for generation}

The utilization of modern VLMs in generative pipelines is increasingly explored. Models like Qwen-Image \cite{wu2025qwenimagetechnicalreport}, UniFusion \cite{li2025unifusionvisionlanguagemodelunified}, FOCUS \cite{yang2025focusunifiedvisionlanguagemodeling}, and UNIMO-G \cite{li2024unimogunifiedimagegeneration} leverage VLMs to encode multimodal inputs, but train the diffusion model, requiring large amounts of domain-specific data.

Conversely, modular approaches align VLM representations to the conditioning spaces of frozen generative models. ThinkDiff \cite{mi2025i}, COSMOS-G \cite{pan2024kosmosggeneratingimagescontext}, and GILL \cite{koh2023generatingimagesmultimodallanguage} utilize aligner networks or self-supervised training to map modern VLM representations to text embedding spaces (e.g., T5 or CLIP). DreamLLM \cite{dong2024dreamllmsynergisticmultimodalcomprehension} avoids alignment to pre-trained spaces entirely, learning a small set of latent query tokens via score distillation to directly condition a frozen model. 
Most directly relevant to our approach is BLIP3-o \cite{chen2025blip3ofamilyfullyopen}, which introduces the use of a diffusion transformer to bridge VLM latents and visual spaces, mapping the trainable latent features produced by a frozen VLM to 64 pooled CLIP-style image embeddings, and providing the latter as input to a frozen image generation decoder. We adopt this generative mapping as the foundation of our work.  

However, such approaches target either text embedding spaces \cite{koh2023generatingimagesmultimodallanguage, pan2024kosmosggeneratingimagescontext, mi2025i}, which lack the dense spatial structure of patch-level visual representations, or compressed image representations \cite{ dong2024dreamllmsynergisticmultimodalcomprehension, chen2025blip3ofamilyfullyopen}, which discard fine-grained spatial information by aggregating patch features into a small number of tokens. Furthermore, both T5 and CLIP-style embeddings benefit from explicit cross-modal alignment during pre-training (e.g., CLIP's contrastive objective), which may simplify the alignment task. In contrast, our work explores mapping to the full embedding space of an image encoder trained exclusively with image-based objectives, without text-to-image alignment. This poses a harder challenge, as the target space is higher-dimensional (e.g., over a thousand patch tokens vs. 64 pooled or 77 text embeddings), and purely visual, lacking explicit cross-modal pre-training.\looseness=-1

\subsection{3D Asset Generation}
Recent 3D asset generation methods have evolved from multiview reconstruction \cite{10.1007/978-3-031-73235-5_1, xu2024instantmeshefficient3dmesh} to native 3D generation \cite{jun2023shapegeneratingconditional3d, 10.1007/978-3-031-73235-5_7,zhang2024gaussiancubestructuredexplicitradianc, chen20253dtopiaxlscalinghighquality3d}. Building on these advancements, TRELLIS \cite{xiang2024structured} introduces a state of the art method for generating 3D assets in multiple output formats, by learning a unified Structured Latent (SLAT) representation. SLAT encodes both geometry and appearance, and is built through a two-stage pipeline, which relies on rectified flow transformers to first generate a sparse 3D structure and then predict appearance details.

However, text-to-3D generation in TRELLIS and similar methods has not reached the fidelity of image-to-3D approaches, exhibiting reduced geometric quality and semantic alignment \cite{xiang2024structured}. Beyond the natural lack of visual detail in text, this is further attributable to the reliance on CLIP text embeddings, which are constrained by a 77-token limit and weak compositional reasoning. Additionally, because image and text conditioning are typically mutually exclusive or performed through separate pipelines, current architectures prevent joint reasoning over multimodal inputs restricting capabilities such as text-guided editing.

\section{Background}

\subsection{Rectified Flow}

We train our generative alignment module using Rectified Flow \cite{liu2022flowstraightfastlearning}, which learns a generative mapping between a standard Gaussian noise source distribution $\pi_0= \mathcal{N}(0, \mathbf{I})$ and the target data distribution $\pi_1$. The generation process is modelled as an Ordinary Differential Equation (ODE) $dx_t = v(x_t, t)dt$, where the vector field $v$ determines the trajectory.

The interpolation path is defined as a linear interpolation between a noise sample $x_0 \sim \pi_0$ and data sample $x_1 \sim \pi_1$:
\begin{equation}
    x_t = t x_1 + (1-t) x_0, \quad t \in [0, 1]
\end{equation}

The constant velocity along this straight path is $u_t = \frac{d x_t}{d t} = x_1 - x_0$. To approximate this vector field, a neural network $v_\theta(x_t, t)$ is trained with the conditional flow matching objective \cite{lipman2023flowmatchinggenerativemodeling}:
\begin{equation}
\mathcal{L}(\theta) = \mathbb{E}_{t \sim \mathcal{U}[0,1], x_0, x_1} \left[ \| v_\theta(x_t, t) - (x_1 - x_0) \|^2 \right]
\end{equation}

\section{Methodology}

Current text-to-3D models must simultaneously resolve visual ambiguity by determining the specific object appearance that corresponds to an abstract text prompt, and perform consistent 3D construction. Relying on global semantic embeddings (e.g., CLIP embeddings) makes this challenge harder, as these may often discard fine-grained details. In contrast, image-to-3D models achieve significantly higher fidelity by conditioning on spatially structured visual features (e.g., obtained from DINOv2), which already provide a strong geometric and appearance prior. GAP3D attempts to bridge this gap by utilizing a modern VLM followed by a generative module to map text prompts into patch-level features, thus transforming the text-to-3D task into a pseudo-image-to-3D problem, and allowing the downstream 3D generative model to focus solely on 3D lifting.\looseness-1

\begin{figure}[ht]
 \vskip 0.2in
   \begin{center}
    
       \centerline{ \includegraphics[width=0.5\textwidth]{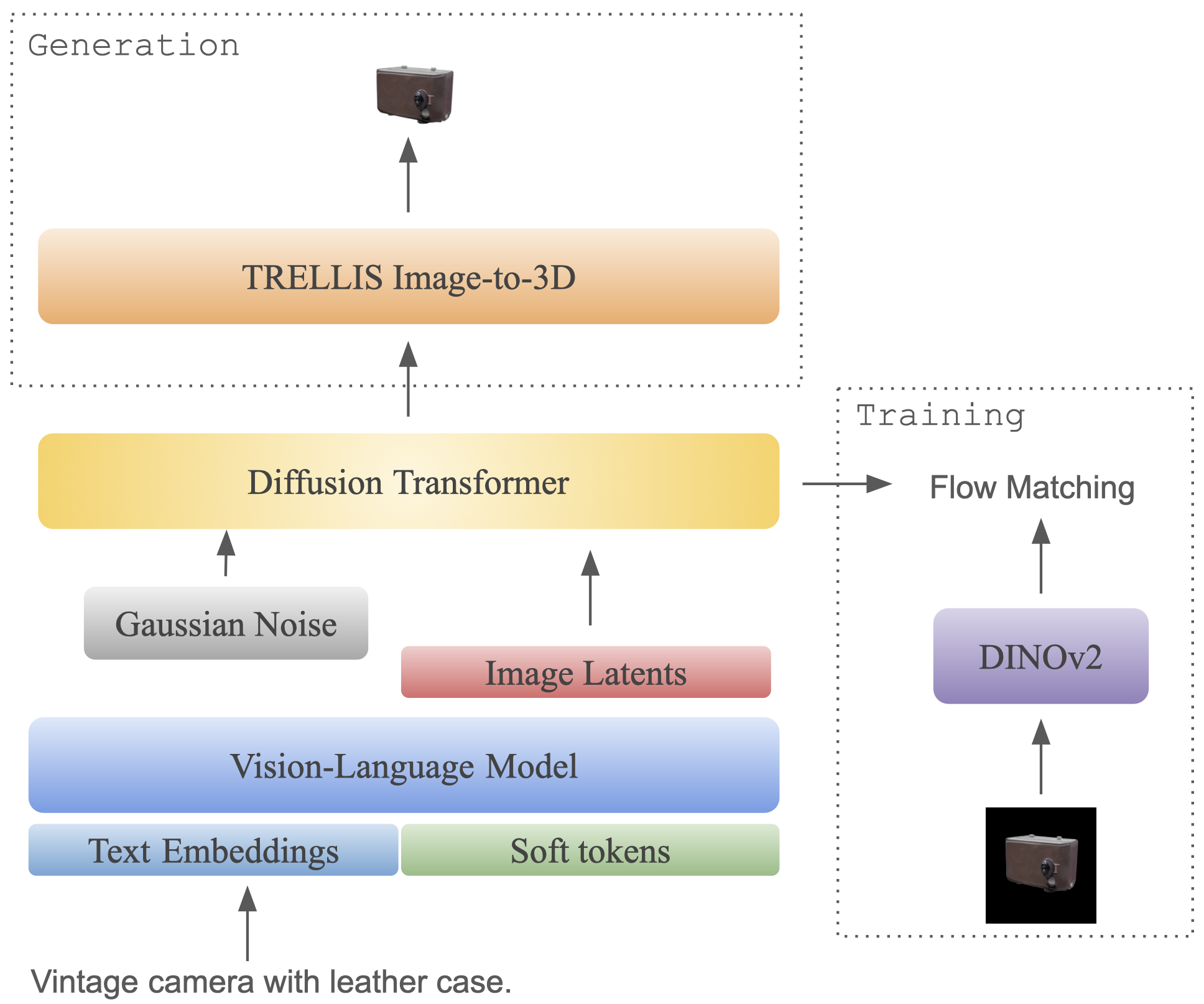}}

    \caption{Training and text-to-3D generation. During training the VLM remains frozen, while the DiT and the soft tokens appended to the VLM's input embeddings are trained via flow matching. During 3D generation, the generated image embeddings condition the frozen TRELLIS image-to-3D model.}
    \label{fig:architecture-ours}
      \end{center}
\end{figure}

\subsection{Diffusion-based Representation Alignment}

Building on BLIP3-o \cite{chen2025blip3ofamilyfullyopen}, we extend generative alignment to a complete image encoder space, mapping jointly to CLS, register, and patch embeddings. An architecture and training overview is provided in Figure \ref{fig:architecture-ours}.

\subsubsection{VLM Encoding}
We utilize a pre-trained VLM to encode the user prompt into a sequence of $s$ latent semantic features, which are then mapped by a diffusion transformer (DiT) to image embeddings in the representation space of our target image encoder. 
We obtain a conditioning signal $\mathcal{C} \in \mathbb{R}^{s \times D_{VLM}}$ by appending $s$ trainable soft tokens to the input text embeddings and extracting their corresponding hidden states. During training, the VLM backbone remains frozen and only the soft tokens are optimized. This preserves the VLM's reasoning capabilities and multimodal priors, maintaining its potential use in understanding tasks, and enabling zero-shot generalization over multimodal prompts (Section~\ref{subsec:Emergent Multimodal Capabilities}).

\subsubsection{Target Representations}
We target the complete feature space of DINOv2~\cite{oquab2024dinov2learningrobustvisual}, a self-supervised vision transformer with robust encoding of both geometric structure and semantic content. For an image resolution of $H \times W$ and patch size $P$, the feature space lies in $\mathbb{R}^{D_{img}}$ and comprises three distinct components:
\begin{itemize}
    \item A dense 2D feature grid $\mathbf{x}^{p} \in \mathbb{R}^{h \times w \times D_{img}}$, where $h=H/P$ and $w=W/P$, encoding appearance and geometry.
    \item A global class token (CLS), $\mathbf{x}^{cls} \in \mathbb{R}^{1 \times D_{img}}$, representing high-level semantic information.
    \item A set of $N_{reg}$ register tokens, $\mathbf{x}^{reg} \in \mathbb{R}^{N_{reg} \times D_{img}}$, which aggregate intermediate global context.
\end{itemize}
We generate all three embedding types jointly to ensure consistency between global semantics and local spatial details.

\subsubsection{Generative Architecture}

We map from the VLM latents $\mathcal{C}$ to the target image feature space using a DiT based on Lumina-Next ~\cite{zhuo2024luminanextmakingluminat2xstronger}, and adapted to fit our setting. The model accepts the noisy states of all three components $\mathbf{x}^{p}_t, \mathbf{x}^{cls}_t, \mathbf{x}^{reg}_t$ as input, and predicts the velocity field along the flow path connecting noise to the target embeddings $\mathbf{x}^{p}_1, \mathbf{x}^{cls}_1, \mathbf{x}^{reg}_1$.

For the $N_{p}=h\times w$ patches, Lumina-Next utilizes a linear patch embedding layer to map the $\sqrt{N_{p}}\times\sqrt{N_{p}}$ grid of patch embeddings to the DiT's hidden representation space. To encode 2D structure, it then employs 2D rotary positional embeddings (RoPE). 

To process all three types of noisy latents, we adapt this embedding layer as follows: For the 2D patch grid $\mathbf{x}^{p}_t$, we utilize Lumina-Next's standard linear patch embedding layer. For the 1D CLS and register tokens ($\mathbf{x}^{cls}_t$ and $\mathbf{x}^{reg}_t$), we introduce specialized linear projection layers to map them to the transformer's hidden dimension.

We then apply hybrid positional embeddings. Specifically, we apply 2D RoPE to the patch grid embedding sequence to enforce the $h \times w$ spatial structure. For CLS and register tokens, since they lack inherent spatial coordinates, we assign identity rotary factors to bypass rotation. To distinguish them from the grid and each other, we add learnable position embeddings to their input projections.

The VLM latents $\mathcal{C}$ are injected into the network via cross-attention layers, guiding the generative process to synthesize visual features aligned with the VLM's semantic interpretation.\looseness-1

\subsection{Training Objective}
We train our diffusion module using rectified flow. We define three synchronized flow paths sharing the same timestep $t$ for the patch grid ($\mathbf{x}^p$), class token ($\mathbf{x}^{cls}$), and register tokens ($\mathbf{x}^{reg}$). For each component, we sample independent Gaussian noise $\mathbf{x}_0$ matching the target's specific dimensions. For the patch component, the noise $\mathbf{x}^p_0 \sim \mathcal{N}(0, \mathbf{I}_{h \times w \times D_{img}})$ is thus sampled as a 2D feature grid, while $\mathbf{x}^{cls}_0$ and $\mathbf{x}^{reg}_0$ are sampled as 1D sequences. The interpolation is applied element-wise:\looseness-1
\begin{equation}
\mathbf{x}^k_t = t \mathbf{x}^k_1 + (1-t) \mathbf{x}^k_0, \quad \text{for } k \in \{p, cls, reg\}
\end{equation}

where $\mathbf{x}^k_1$ represents the ground-truth DINOv2 embeddings. The diffusion module $v_\theta$ accepts the set of all noisy states $\{\mathbf{x}^p_t, \mathbf{x}^{cls}_t, \mathbf{x}^{reg}_t\}$ and, conditioned on $\mathcal{C}$, simultaneously predicts the velocity fields for all three components. This allows global (CLS, registers) and local (patches) components to interact during the denoising process through self-attention, ensuring consistency. The final objective is a weighted sum of the flow matching losses:
\begin{equation}
\mathbf{x}_t := (\mathbf{x}^p_t, \mathbf{x}^{cls}_t, \mathbf{x}^{reg}_t)
\end{equation}
\begin{equation}
\mathcal{L}(\theta) = \mathbb{E}_{t, \mathbf{x}_0, \mathbf{x}_1} \left[
\sum_{k} \lambda_k \left\|
\mathbf{v}^k_\theta(\mathbf{x}_t, t, \mathcal{C})
- (\mathbf{x}^k_1 - \mathbf{x}^k_0)
\right\|^2
\right]
\end{equation}
where $\mathbf{v}^k_\theta$ denotes the model's prediction for the velocity field component $k$, and $\lambda_k$ are balancing weights.

\subsection{Downstream Integration with TRELLIS} \label{sec: trellis integration}

To evaluate GAP3D in the context of 3D asset generation, we utilize TRELLIS as our downstream generative backbone. TRELLIS natively supports both image-to-3D (via DINOv2) and text-to-3D (via CLIP) generation through two distinct pipelines. We utilize the image-to-3D pipeline, but bypass its image encoder. Instead, we encode a text prompt through our VLM and diffusion module, and inject the generated dense visual features directly as the conditioning embeddings for the frozen TRELLIS model. This enables text-to-3D generation, while utilizing the spatially structured image conditioning branch. A comparative architecture diagram is provided in Appendix \ref{appendix:architecture}.

\subsection{Domain Adaptation}
We pre-train our VLM latents and diffusion module on large-scale paired image-text data featuring diverse scenes with variable numbers of objects and complex backgrounds. However, these natural images differ significantly from the 3D asset renderings our downstream 3D generative model (TRELLIS) was trained on, which typically feature clean, object-centric views with transparent or black backgrounds and uniform lighting.  This introduces a distribution shift to which the frozen 3D model is sensitive, leading to reduced visual quality and artifacts in the generation. 

To address this, we adopt a two-stage training curriculum. We first pre-train the diffusion module and soft-tokens on diverse, large-scale image-text pairs, to establish a robust mapping between semantic concepts and visual features in the target image encoder space, leveraging the large number of available general domain image - text pairs. This stage teaches the model a generalizable mapping to the image encoding space, by exposing it to a variety of semantic concepts, visual features, geometry, textures and more. Subsequently, we conduct fine-tuning on a smaller number of images derived from domain-specific 3D renderings, matching the distribution of conditioning images utilized in the 3D model's pre-training phase.

\begin{table*}[t]
\caption{Cosine similarity (COS), Mean Squared Error (MSE) and Norm ratio (Norm R.) between target and generated embeddings for GAP3D and BLIP3-o baseline. For patch embeddings the average across all tokens is presented.}
\label{cosine_sim_general}
\begin{flushleft}
 \begin{small}
  \begin{sc}
\begin{adjustbox}{width=\textwidth}
\setlength{\tabcolsep}{4pt}
\begin{tabular}{l|ccc|ccc|ccc|ccc}
\toprule
& \multicolumn{6}{c|}{COCO} & \multicolumn{6}{c}{Toys4K} \\
& \multicolumn{3}{c}{Patch} & \multicolumn{3}{c|}{CLS}
& \multicolumn{3}{c}{Patch} & \multicolumn{3}{c}{CLS} \\ 
& Cos $\uparrow$ & MSE $\downarrow$ & Norm r.$\rightarrow 1$
& Cos $\uparrow$ & MSE $\downarrow$ & Norm r.$\rightarrow 1$
& Cos $\uparrow$ & MSE $\downarrow$ & Norm r.$\rightarrow 1$
& Cos $\uparrow$ & MSE $\downarrow$ & Norm r.$\rightarrow 1$ \\
\midrule
BLIP3-o
& 0.36 & 7.33 & 1.33
& --    & --    & --
& 0.45 & 7.33 & 1.36
& --    & --    & -- \\
GAP3D (Pre-trained)
& \textbf{0.43} & \textbf{1.10} & 0.97
& \textbf{0.53} & \textbf{0.92} & \textbf{0.98}
& 0.38 & 1.20 & 0.98
& 0.58 & 0.82 & \textbf{0.98} \\
GAP3D (Fine-tuned)
& 0.28 & 1.44 & \textbf{1.00}
& 0.36 & 1.22 & 0.94
& \textbf{0.67} & \textbf{0.67} & \textbf{1.00}
& \textbf{0.69} & \textbf{0.60} & 0.97 \\
\bottomrule
\end{tabular}
 \end{adjustbox}
 \end{sc}
 \end{small}
\end{flushleft}
\end{table*}

\section{Experiments}\label{sec:experiments}

Our experimental evaluation assesses the effectiveness of mapping VLM representations to dense visual spaces and their utility for downstream generation. Specifically, we investigate: (Q1) Alignment Feasibility (Sec. \ref{subsec:eval_of_repr_alignement}, \ref{subsec:text-to-image-retrieval}): Whether VLM latents can be mapped to the high-dimensional, patch-level embedding space of a pre-trained image encoder while preserving semantic content. (Q2) Geometric and Visual Fidelity (Sec. \ref{subsec:text-to-3D}): The extent to which the dense embeddings capture fine-grained geometry and structural detail, evaluated via downstream text-to-3D generation. (Q3) Multimodal Transfer (Sec. \ref{subsec:Emergent Multimodal Capabilities}): If the inherent multimodal representations of the frozen VLM can enable multimodal prompting zero-shot, despite our text-only training prompts. Finally, in Appendix \ref{appendix:prompting ablation}, we explore (Q4) Domain Adaptation: The necessity of fine-tuning to bridge the distribution shift between natural images and synthetic 3D renderings.

\subsection{Experimental Setup}

\subsubsection{Datasets}

We pre-train our model on the public BLIP3-o dataset, comprising approximately 31 million image-text pairs. For domain-specific fine-tuning, we curate a dataset of around 60k images by rendering a single canonical view per asset from a random subset of Objaverse-XL \cite{deitke2023objaversexluniverse10m3d}. To evaluate the general-domain embedding alignment of our method, we use the MS-COCO \cite{lin2015microsoftcococommonobjects} validation set, which consists of 5000 images. For 3D-specific alignment and downstream 3D asset generation, we utilize Toys4K \cite{stojanov2021usingshapecategorizelowshot}.  More information on data augmentation strategies and the generation of 3D asset - originating images can be found in Appendix \ref{appendix:data-generation}.

\subsubsection{Implementation Details}
We use the "dinov2-vitl14-reg" encoder as our target, processing $518 \times 518$ images into a $37 \times 37$ patch grid ($N_p = 1369$), one CLS embedding and 4 register embeddings ($N_{reg} = 4)$  with dimension $D_{img} = 1024$. We utilize Qwen2.5-VL-3B-Instruct as our VLM. Our trainable generative alignment module is a DiT based on the Lumina-Next architecture~\cite{zhuo2024luminanextmakingluminat2xstronger}. This results in approximately 4.75 billion parameters for our VLM and diffusion module, out of which approximately 1.36 billion are trainable.  Further training details are provided in Appendix \ref{appendix:hyperparameters}.

\begin{table*}[t]
\caption{Top-K recall (\%), on MS-COCO and Toys4K, for zero-shot text-to-image retrieval. Results using the average of patch embeddings and the CLS embedding are presented for our models, while only the pooled EVA-CLIP patch embeddings are utilized for BLIP3-o.}
\label{tab:image_retrieval_metrics}
\begin{center}

\begin{small}
      \begin{sc}
      \begin{adjustbox}{width=\textwidth}
\begin{tabular}{l|ccc|ccc|ccc|ccc}
\toprule
& \multicolumn{6}{c|}{MS-COCO} & \multicolumn{6}{c}{Toys4K} \\
& \multicolumn{3}{c}{Pooled-Patch} & \multicolumn{3}{c|}{CLS}
& \multicolumn{3}{c}{Pooled-Patch} & \multicolumn{3}{c}{CLS} \\
& R@1 & R@5 & R@10 & R@1 & R@5 & R@10
& R@1 & R@5 & R@10 & R@1 & R@5 & R@10 \\
\midrule
BLIP3-o
& \textbf{7.17} & \textbf{37.49} & \textbf{70.50}
& --   & --    & --
& \textbf{7.42} & \textbf{42.52} & \textbf{80.06}
& --   & --    & -- \\
GAP3D (Pre-trained)
& 4.98 & 24.02 & 44.98
& \textbf{4.47} & \textbf{23.06} & \textbf{44.18}
& 1.54 & 9.09  & 16.86
& 5.53 & 28.65 & 53.62 \\
GAP3D (Fine-tuned)
& 0.27 & 1.77 & 4.08
& 1.94 & 10.44 & 19.73
& 2.96 & 14.81 & 27.48
& \textbf{6.64} & \textbf{33.55} & \textbf{63.43} \\
\bottomrule
\end{tabular}
 \end{adjustbox}
\end{sc}
\end{small}
\end{center}
\end{table*}

\begin{table*}[t]
\caption{Quantitative results for 3D asset generation for TRELLIS image-to-3D and text-to-3D versions, and for our pre-trained and fine-tuned models. KD is reported $\times 100$.}
\label{tab:toys4k_3d_metrics}
\begin{center}
\begin{small}
    \begin{sc}
\begin{adjustbox}{width=\textwidth}
\setlength{\tabcolsep}{6pt}
\begin{tabular}{l|cccccc}
\toprule
& FD (Inception) $\downarrow$
& KD (Inception) $\downarrow$
& FD (DINOv2) $\downarrow$
& KD (DINOv2) $\downarrow$
& FD (PointNet++) $\downarrow$
& CLIP $\uparrow$ \\
\midrule
\multicolumn{7}{c}{\textbf{Image $\rightarrow$ 3D}} \\
\midrule
TRELLIS 
& \textbf{12.27} & \textbf{0.04} & \textbf{106.97} & \textbf{2.21} & \textbf{25.86} & \textbf{91.62} \\
\midrule
\multicolumn{7}{c}{\textbf{Text $\rightarrow$ 3D}} \\
\midrule
TRELLIS 
& 21.86 & 0.08 & 287.64 & 5.60 & 26.54 & 30.52 \\
GAP3D (Pre-trained)
& 41.89 & 0.88 & 634.57 & 52.71 & 90.89 & 27.51 \\
GAP3D (Fine-tuned)
& 24.20 & 0.16 & 329.90 & 10.65 & 44.37 & 29.26 \\
\bottomrule
\end{tabular}
\end{adjustbox}
 \end{sc}
\end{small}
\end{center}
\end{table*}

\begin{figure}[ht]
\centering
\setlength{\tabcolsep}{2pt} 
\resizebox{1\linewidth}{!}{
\begin{tabular}{c c c c c}
\toprule
\textbf{GT} & \textbf{Pos 1} & \textbf{Pos 2} & \textbf{Pos 3} & \textbf{Pos 4}  \\
\midrule

\includegraphics[width=0.16\linewidth]{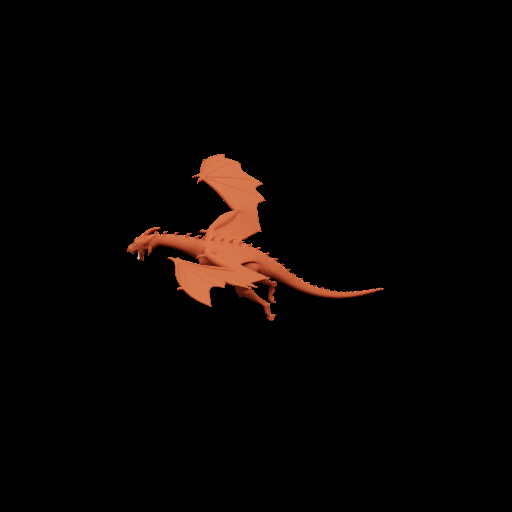} &
\includegraphics[width=0.16\linewidth]{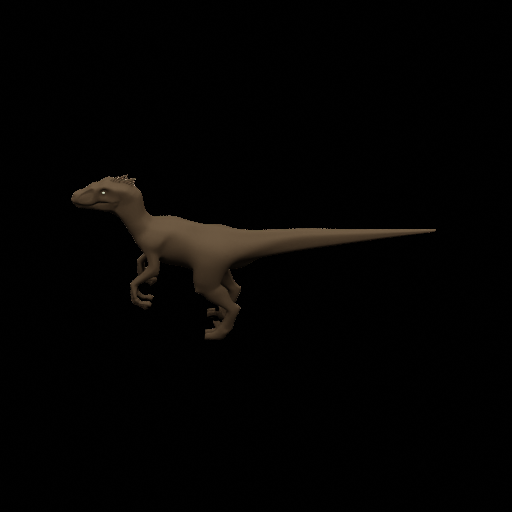} &
\includegraphics[width=0.16\linewidth]{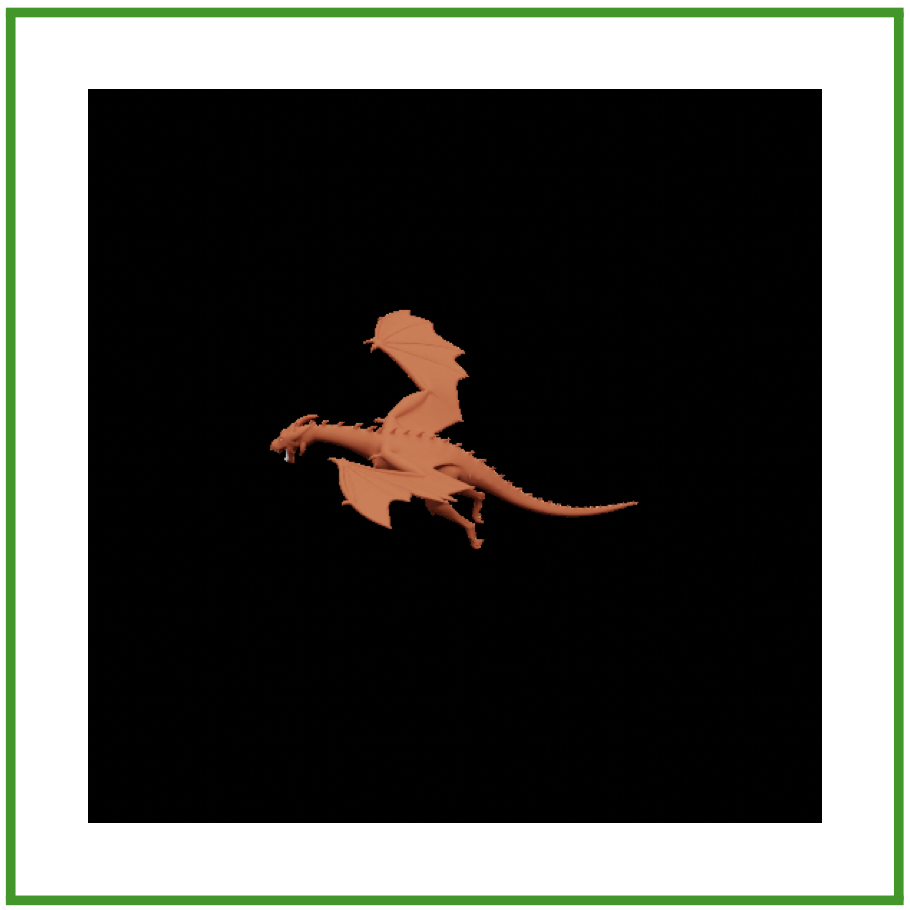} &
\includegraphics[width=0.16\linewidth]{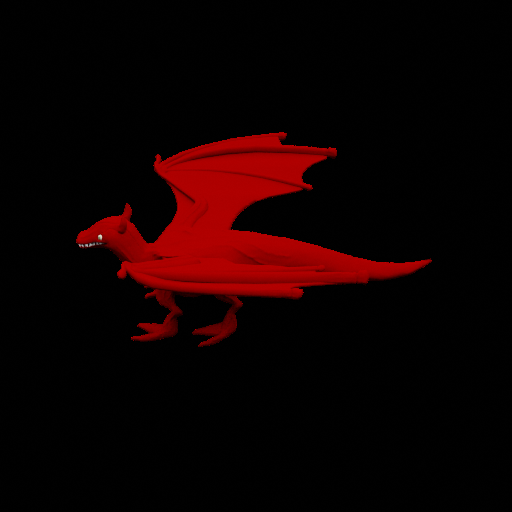} &
\includegraphics[width=0.16\linewidth]{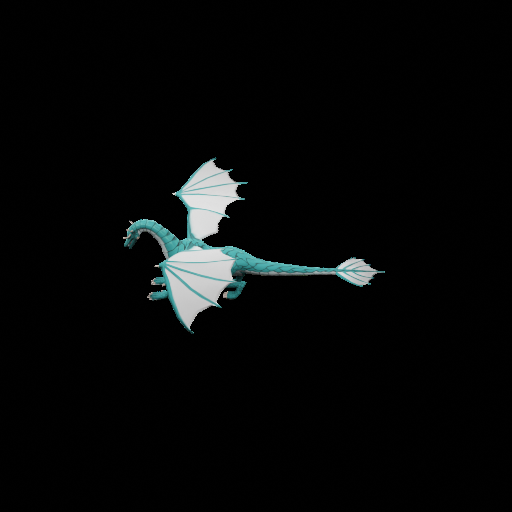} \\
\multicolumn{5}{p{0.9\linewidth}}{%
\small
\emph{Example 1:}
"Uniform red dragon flying, with elongated head, small horns, ... long ... tail with slight upward curve."
}\\
\addlinespace[4pt]

\includegraphics[width=0.16\linewidth]{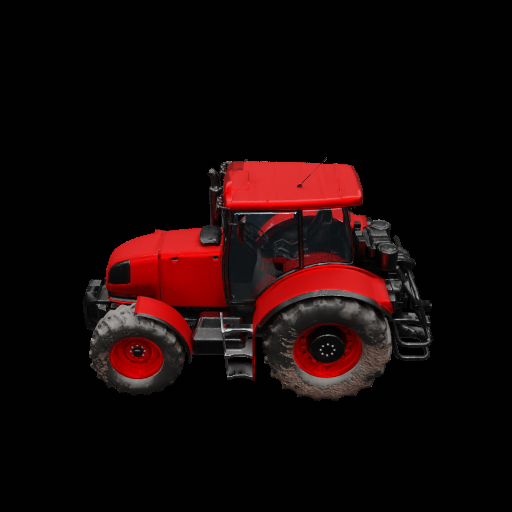} &
\includegraphics[width=0.16\linewidth]{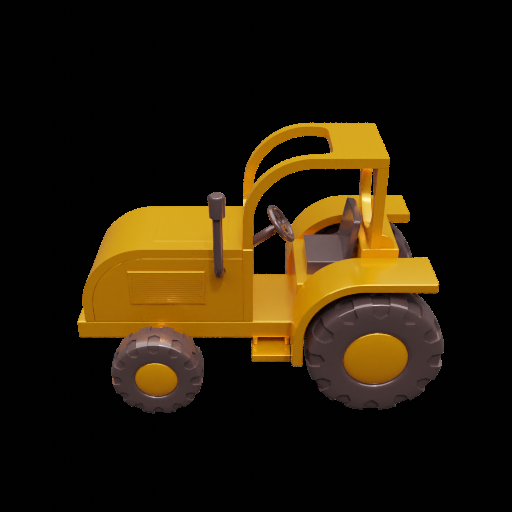} &
\includegraphics[width=0.16\linewidth]{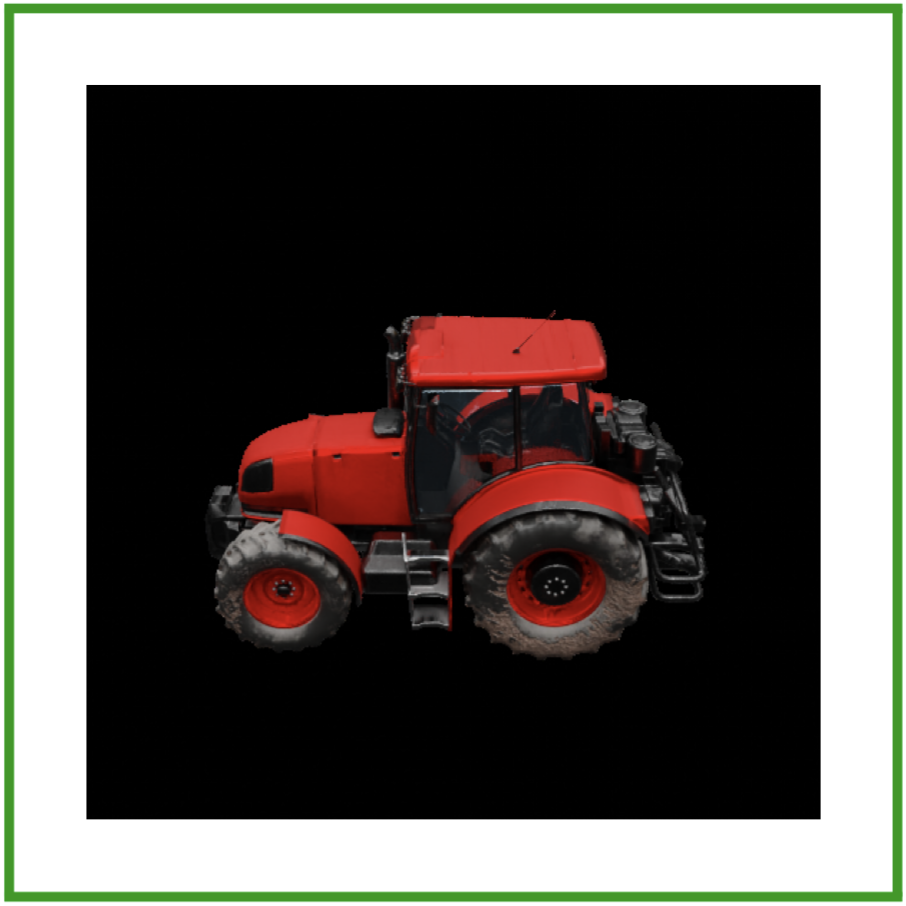} &
\includegraphics[width=0.16\linewidth]{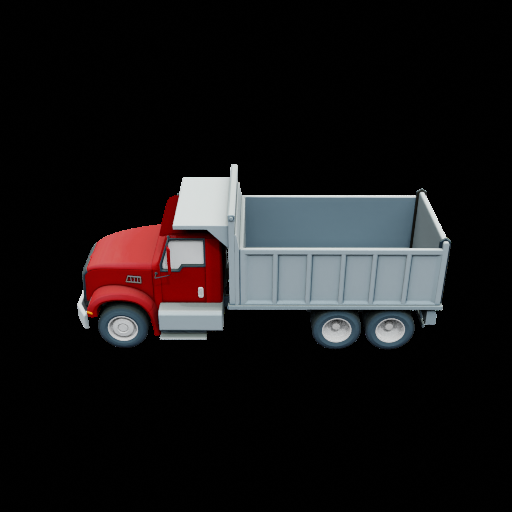} &
\includegraphics[width=0.16\linewidth]{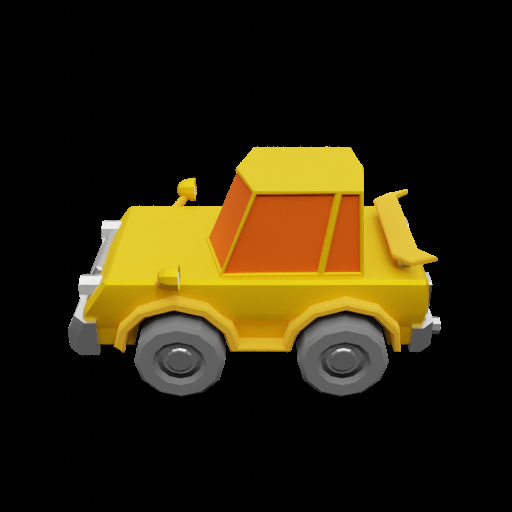} \\

\multicolumn{5}{p{0.9\linewidth}}{%
\small
\emph{Example 2:}
"A red tractor with a cuboid cab, rounded rectangular hood, and ... attachment points, and rear lights."
}\\
\addlinespace[4pt]

\includegraphics[width=0.16\linewidth]{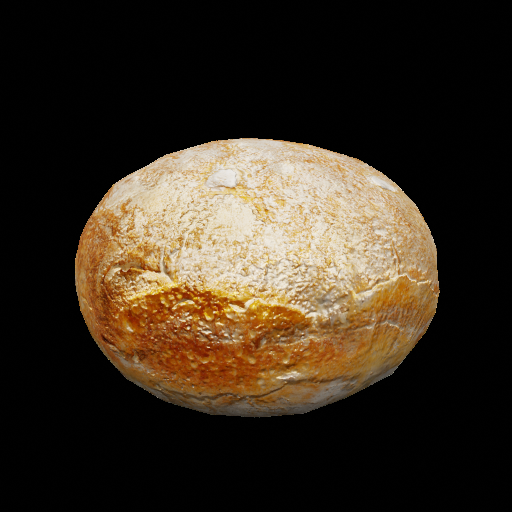} &
\includegraphics[width=0.16\linewidth]{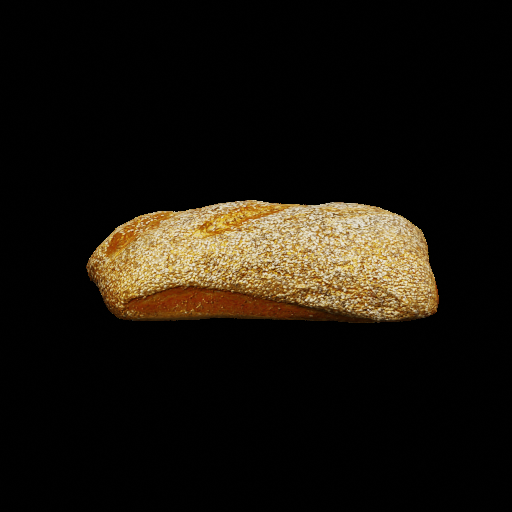} &
\includegraphics[width=0.16\linewidth]{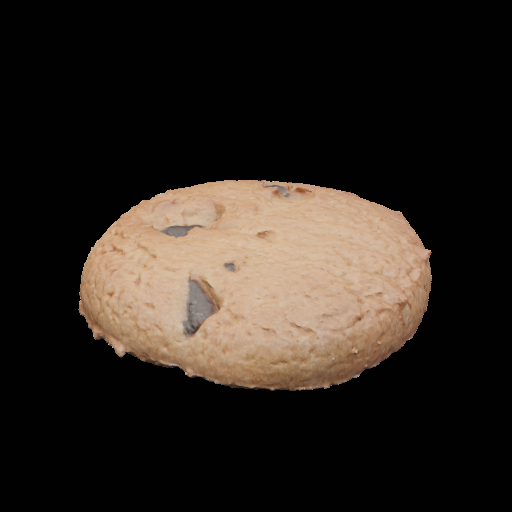} &
\includegraphics[width=0.16\linewidth]{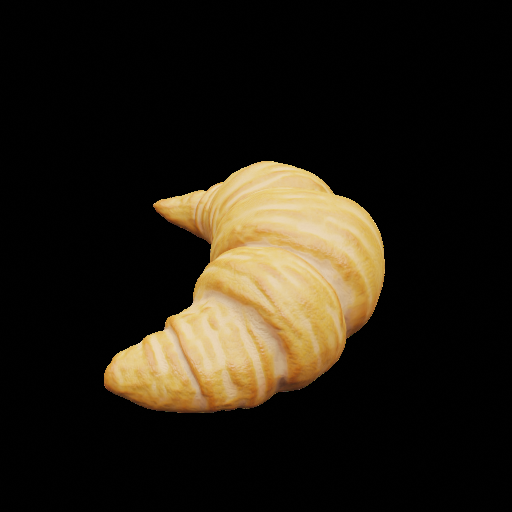} &
\includegraphics[width=0.16\linewidth]{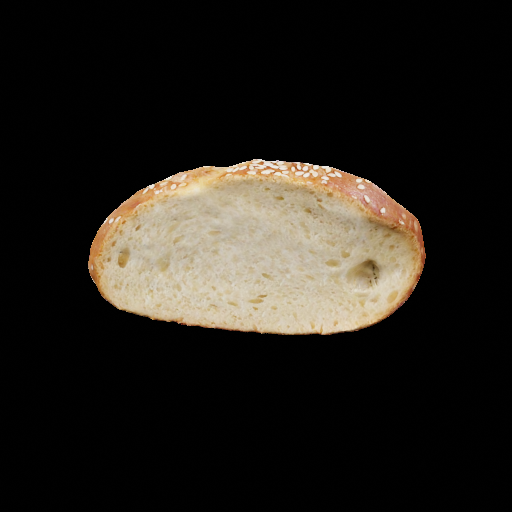} \\

\multicolumn{5}{p{0.9\linewidth}}{%
\small
\emph{Example 3:}
"Oval-shaped, golden-brown artisanal bread loaf with a crackled, ... crust, ... brown underside."
}\\

\end{tabular}
}
\caption{Qualitative image retrieval results using our fine-tuned model. Each row shows a query (GT) and its top-4 retrieved images. When correctly retrieved, the image is highlighted.}
\label{fig:retrieval_grid}
\end{figure}

\subsection{Evaluation of Representation Alignment} \label{subsec:eval_of_repr_alignement}

To quantify alignment quality, we compute the cosine similarity, mean squared error (MSE), and norm ratio between the ground-truth DINOv2 embeddings and those generated by GAP3D. We evaluate two versions of our model: one pre-trained solely on general image-text pairs, and one further fine-tuned on 3D asset renderings. Table~\ref{cosine_sim_general} presents results for both general natural images (MS-COCO) and 3D asset renderings (Toys4K). We additionally include metrics for the embeddings generated by the pre-trained BLIP3-o model. While not directly comparable due to its pooled EVA-CLIP target space, this provides a reference for alignment quality in a compressed versus full, dense visual space.

Comparing our pre-trained and fine-tuned models reveals a significant distribution shift. The pre-trained model achieves higher alignment on MS-COCO, reflecting its diverse training data, whereas the fine-tuned model substantially improves alignment on Toys4K, showcasing an effective adaptation to the embedding distribution of the clean, object-centric 3D renders. However, this comes at the cost of catastrophic forgetting on the now out-of-distribution MS-COCO natural images. This trade-off could likely be mitigated by mixing general and 3D domain data during fine-tuning.

Across all settings, the CLS embeddings achieve better alignment than patch embeddings. This is expected given our loss formulation, which assigns a significantly higher per-token weight to global tokens (CLS, registers). Generating a global image descriptor is also inherently less ambiguous than hallucinating dense, patch-aligned spatial details from text, making patch embedding alignment a fundamentally more ill-defined task when starting from text prompts.

Overall, despite the DINOv2 target space's significantly higher dimensionality (1,369 patch tokens vs. 64 for BLIP3-o), Table~\ref{cosine_sim_general} indicates that our model establishes a meaningful text-to-visual mapping. Our generated embeddings yield norm ratios closer to 1.0 than BLIP3-o's EVA-CLIP embeddings, and comparable cosine similarity. This is despite the fact that the evaluated BLIP3-o model utilizes a larger 7B-parameter backbone (versus our 3B model) and was trained on an additional 30 million proprietary images. Finally, the higher patch cosine similarity (0.67) achieved on Toys4K indicates that the alignment task is learnable, suggesting that longer pre-training or refined dataset construction could possibly yield an even stronger, generalizable mapping.

\begin{figure*}[ht]
\vskip 0.2in
  \begin{center}
\setlength{\tabcolsep}{2pt} 
\resizebox{0.87\linewidth}{!}{
\begin{tabular}{c c c c c c}
\toprule
\textbf{TRELLIS (G.)} & \textbf{TRELLIS (M.)} & \textbf{Fine-tuned (G.)} & \textbf{Fine-tuned (M.)} & \textbf{Pre-trained (G.)} & \textbf{Pre-trained (M.)} \\
\midrule

\includegraphics[width=0.16\linewidth]{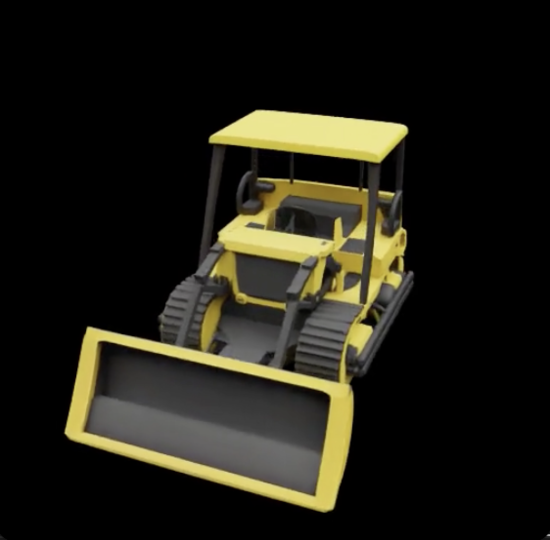} &
\includegraphics[width=0.16\linewidth]{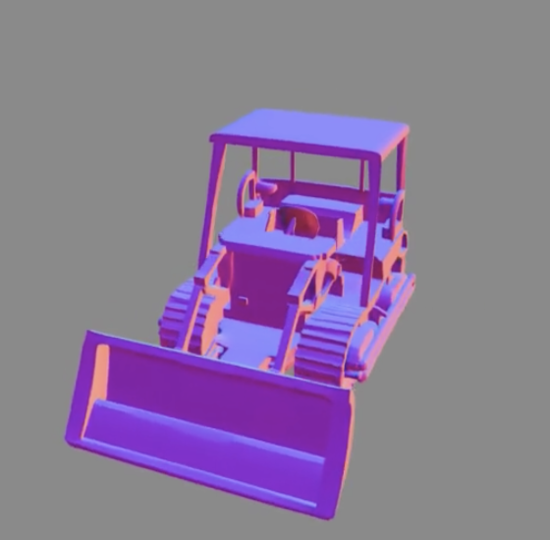} &
\includegraphics[width=0.16\linewidth]{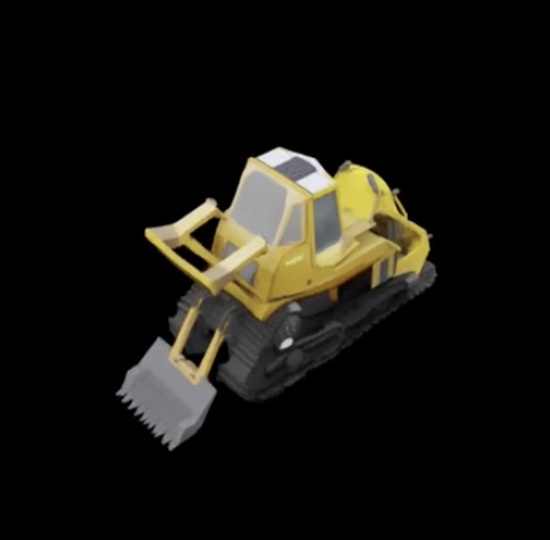} &
\includegraphics[width=0.16\linewidth]{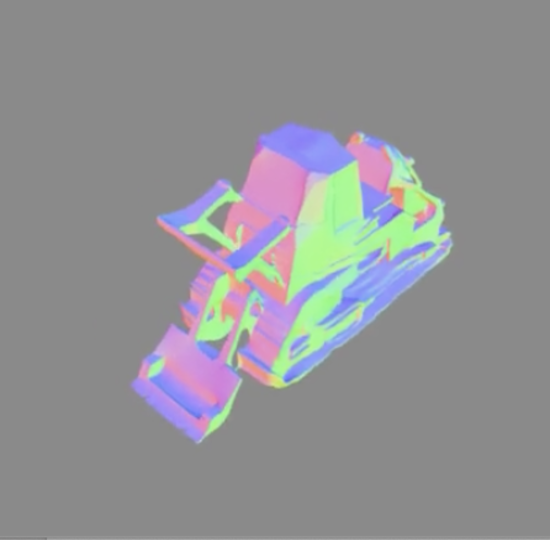} &
\includegraphics[width=0.16\linewidth]{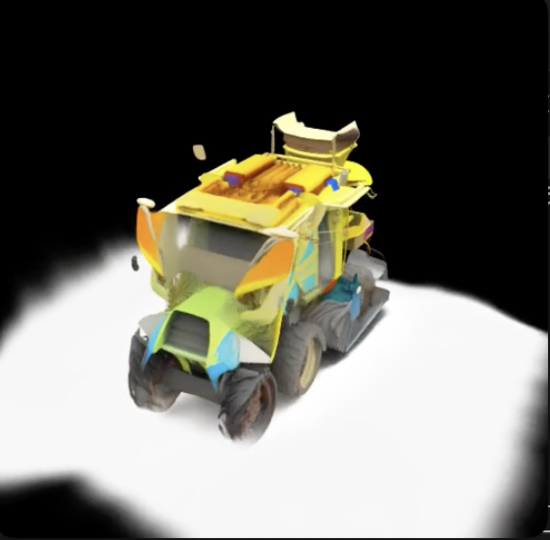} &
\includegraphics[width=0.16\linewidth]{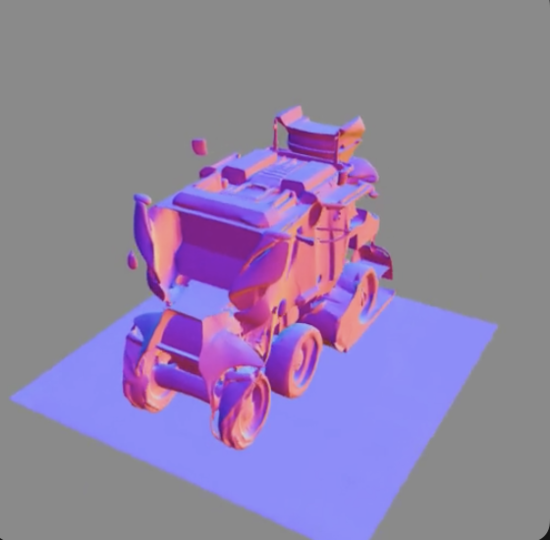} \\
\multicolumn{6}{p{0.95\linewidth}}{%
\small
\emph{Example 1:}
"Yellow and black bulldozer with movable front blade."
}\\

\addlinespace[4pt]

\includegraphics[width=0.16\linewidth]{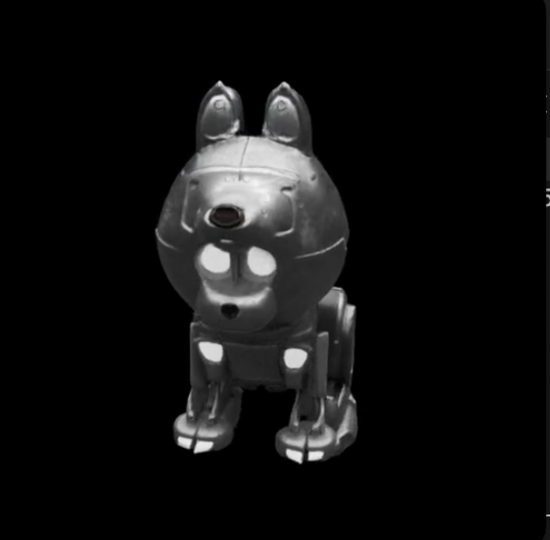} &
\includegraphics[width=0.16\linewidth]{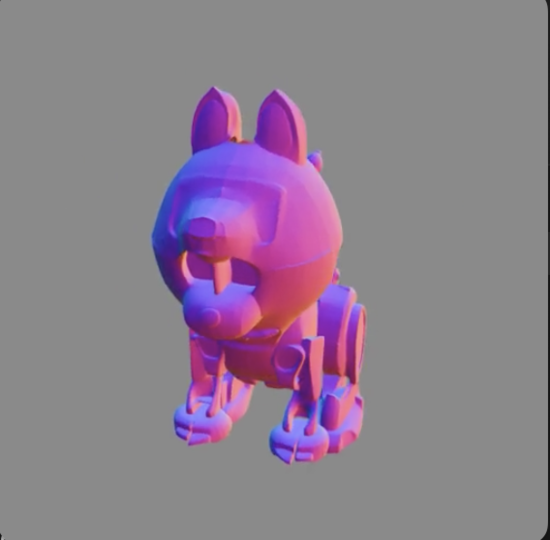} &
\includegraphics[width=0.16\linewidth]{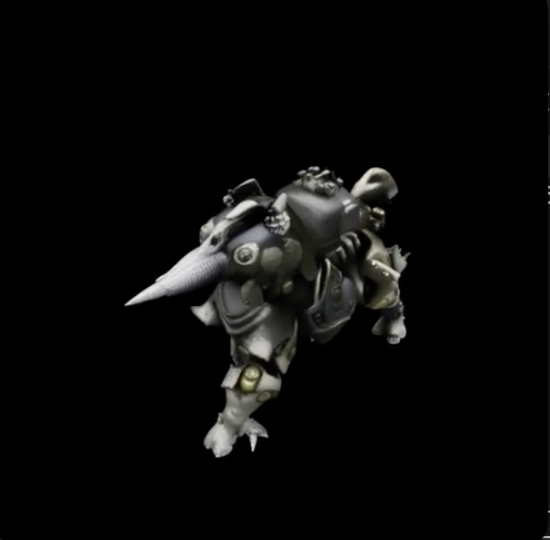} &
\includegraphics[width=0.16\linewidth]{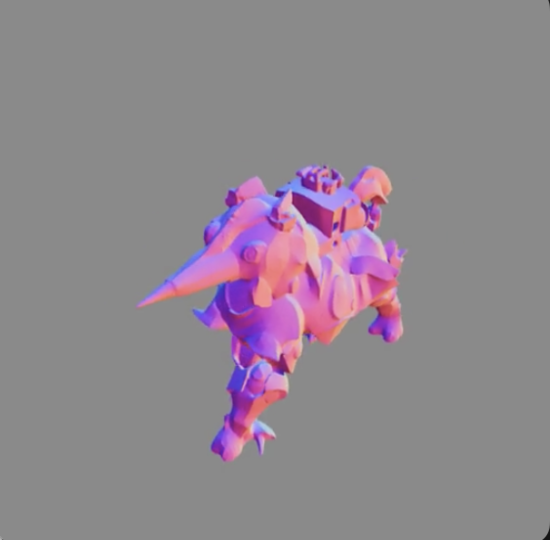} &
\includegraphics[width=0.16\linewidth]{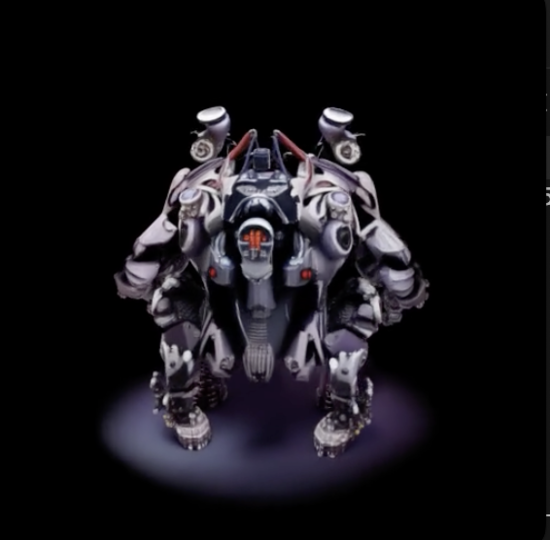} &
\includegraphics[width=0.16\linewidth]{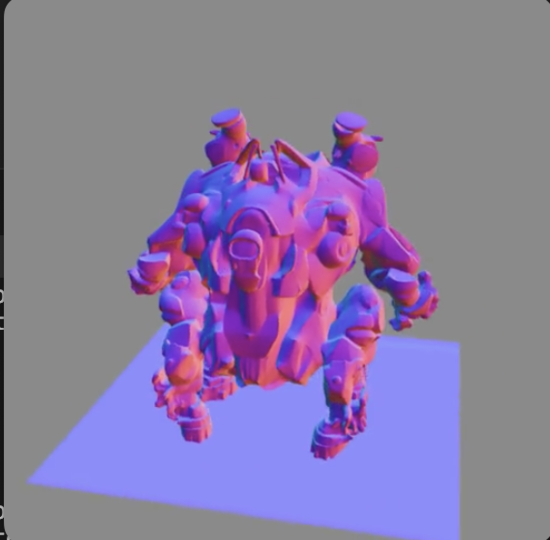} \\

\multicolumn{6}{p{0.9\linewidth}}{%
\small
\emph{Example 2:}
"Metallic dog-like robot with articulated legs and futuristic design elements."
}\\

\addlinespace[4pt]

\includegraphics[width=0.16\linewidth]{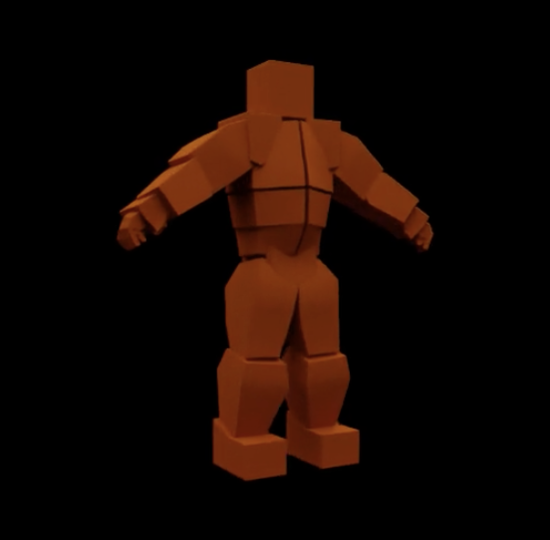} &
\includegraphics[width=0.16\linewidth]{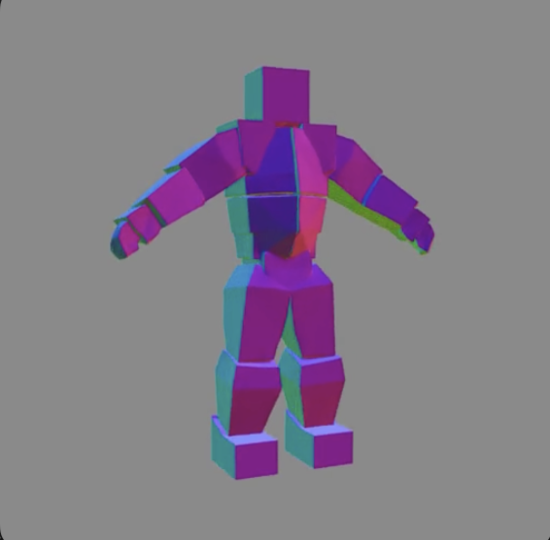} &
\includegraphics[width=0.16\linewidth]{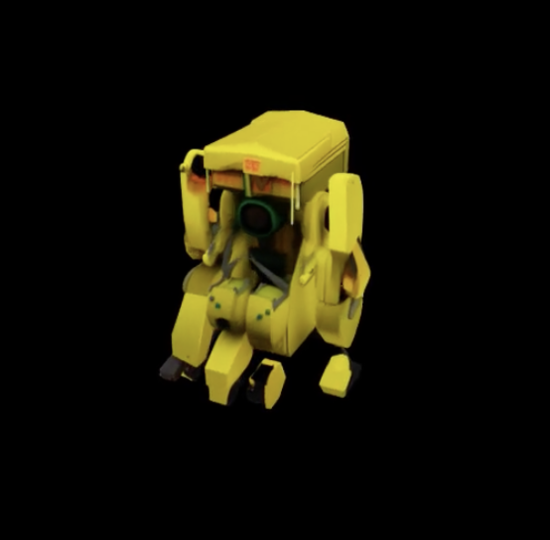} &
\includegraphics[width=0.16\linewidth]{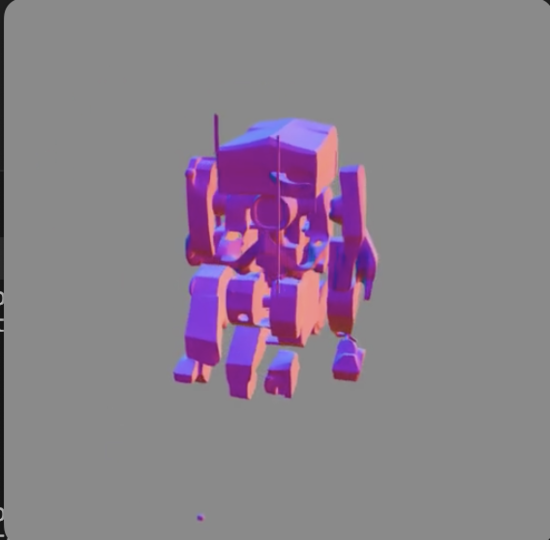} &
\includegraphics[width=0.16\linewidth]{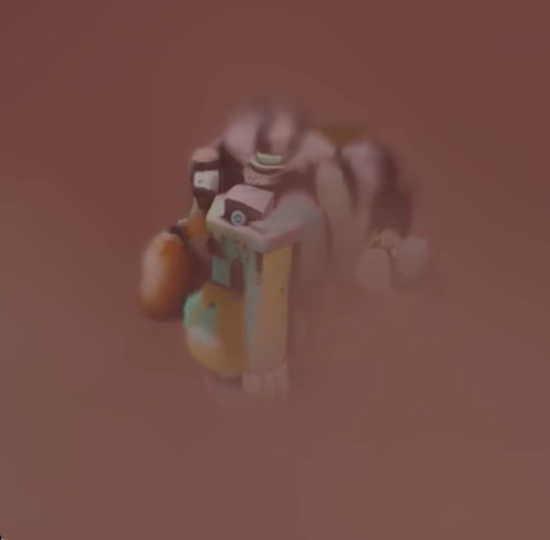} &
\includegraphics[width=0.16\linewidth]{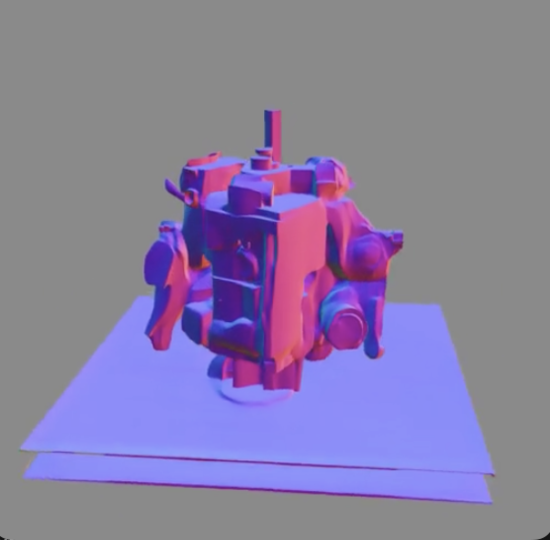} \\

\multicolumn{6}{p{0.9\linewidth}}{%
\small
\emph{Example 3:}
"Blocky, orange and teal robot with articulated limbs."
}\\

\includegraphics[width=0.16\linewidth]{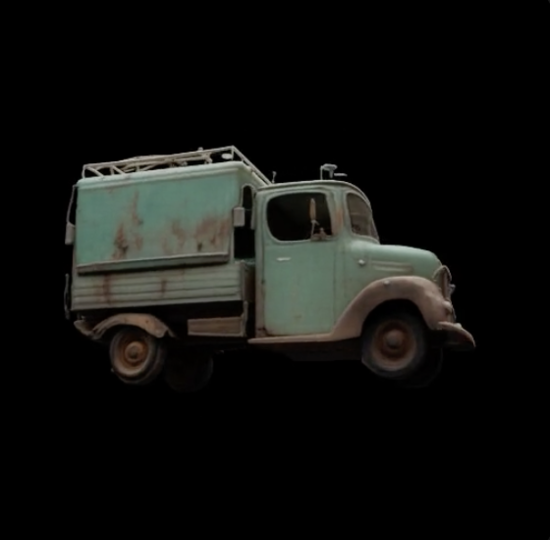} &
\includegraphics[width=0.16\linewidth]{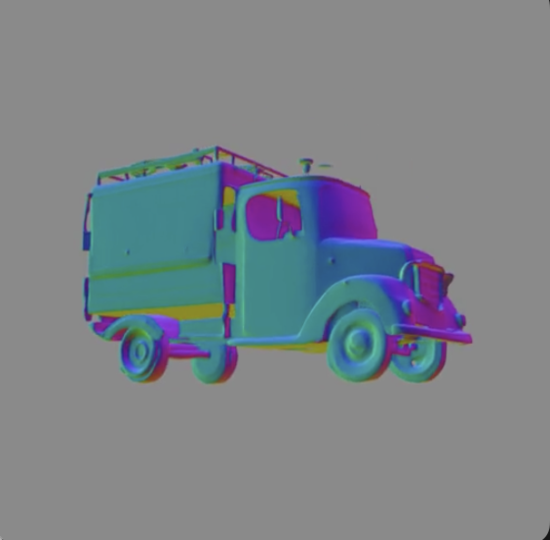} &
\includegraphics[width=0.16\linewidth]{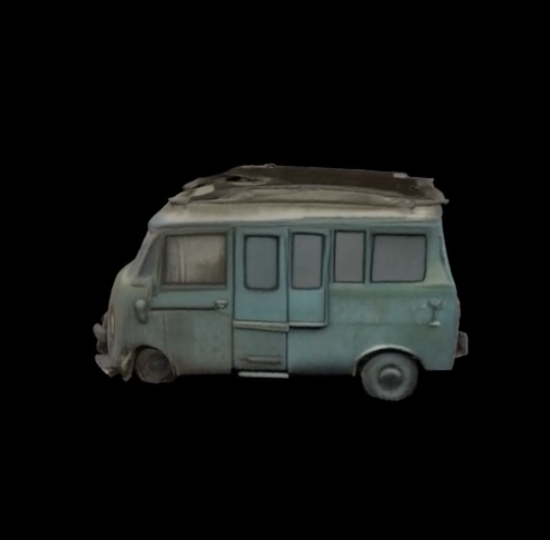} &
\includegraphics[width=0.16\linewidth]{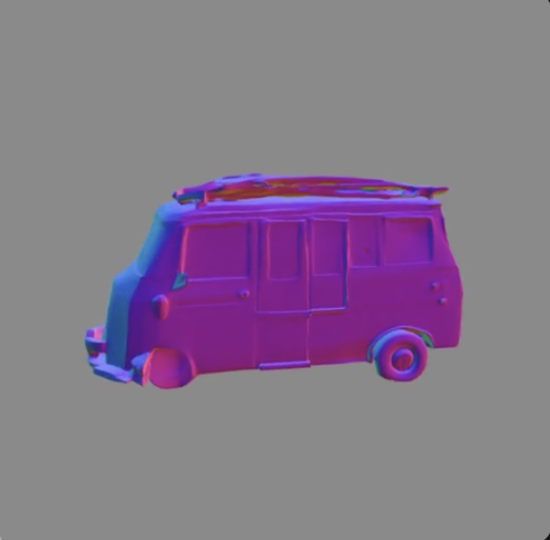} &
\includegraphics[width=0.16\linewidth]{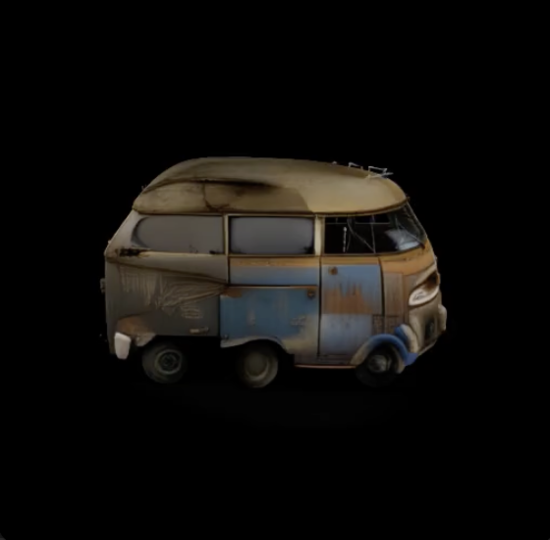} &
\includegraphics[width=0.16\linewidth]{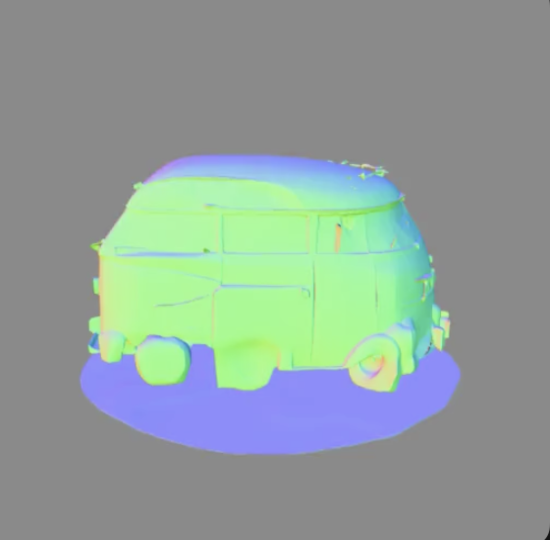} \\

\multicolumn{6}{p{0.98\linewidth}}{%
\small
\emph{Example 4:}
"A weather-worn vintage delivery van with a boxy shape, a rusted faded green finish, square windows, rusty roof rack."
}\\
\end{tabular}
}
\end{center}
\caption{3D asset generation examples for TRELLIS text-to-3D baseline, our pre-trained, and our fine-tuned model. The generated 3D Gaussian (G.) and Mesh (M.) are presented for each asset.}
\label{fig:3d_asset_grid}
\end{figure*}

\subsection{Zero-shot Text-to-Image retrieval} \label{subsec:text-to-image-retrieval}
To assess the level of semantic detail captured by our generated embeddings, we extend our evaluation to zero-shot text-to-image retrieval. For this purpose, we use cosine similarity to query a database of ground-truth DINOv2 embeddings using either the generated CLS embedding, or averaged patch embeddings derived by providing the corresponding image captions as input to our VLM and DiT. In addition to our pre-trained and fine-tuned models, we report results for BLIP3-o (using pooled patch embeddings, as it does not generate a CLS EVA-CLIP embedding). This comparison is again not entirely fair, due to differences in base model sizes and target embedding spaces, but rather serves primarily as a proxy for comparing performance when mapping to a full, visual target space versus a pooled text-aligned space.

As shown in Table~\ref{tab:image_retrieval_metrics}, we again observe a clear distribution shift, with the pre-trained model performing better on general MS-COCO data, and the fine-tuned model prevailing on Toys4K. In most cases, the CLS embedding demonstrates superior retrieval performance, reflecting its better alignment compared to the patch embeddings. BLIP3-o achieves notably higher recall on both datasets. This likely reflects the lower difficulty of the target embedding space alignment task, but may also be attributed to BLIP3-o's larger training dataset and model size, or potentially suboptimal loss weighting and hyperparameter choice in our training.

 Visually (Figure~\ref{fig:retrieval_grid}), the fine-tuned model retrieves the correct object class or semantically similar items (e.g., bread-croissant, dragon-dinosaur), demonstrating that our generative mapping successfully encodes high-level semantics. However, lower-level visual details like specific colours or shapes, are not fully captured. This may indicate a noisy embedding space or suggests that the VLM does not explicitly provide such fine-grained detail in its image latents. The latter could be mitigated in future work by increasing the number of image latents to enhance the capacity for encoding fine detail, or by incorporating image reconstruction pre-training objectives to strengthen low-level detail encoding.

\subsection{Text-to-3D Generation} \label{subsec:text-to-3D} 

To evaluate downstream utility, we condition TRELLIS on our generated DINOv2 embeddings (see Section \ref{sec: trellis integration}).
Following Xiang et al. \yrcite{xiang2024structured}, we measure visual and geometric fidelity using Fréchet Distance (FD)~\cite{heusel2018ganstrainedtimescaleupdate} and Kernel Distance (KD)~\cite{bińkowski2021demystifyingmmdgans} with DINOv2, Inception-v3 \cite{szegedy2015rethinkinginceptionarchitecturecomputer} and PointNet++ \cite{qi2017pointnetdeephierarchicalfeature} as feature extractors, and evaluate semantic text-to-3D alignment via CLIP Score~\cite{radford2021learningtransferablevisualmodels}. Appendix~\ref{appendix:evaluation} provides a detailed explanation of our evaluation protocol. We compare our pre-trained and fine-tuned models with the original TRELLIS pipelines (Image-to-3D and Text-to-3D). 

As our approach utilizes pre-trained TRELLIS fully frozen, the final generation quality is determined solely by the quality of our conditioning signal (generated DINOv2 embeddings) and TRELLIS's original capabilities. Thus, a direct comparison with the pre-trained TRELLIS baselines is sufficient to judge the fidelity of our alignment module, and we do not compare with different 3D generative architectures. Following Xiang et al. \cite{xiang2024structured}, we perform all evaluations on the Toys4K dataset.\looseness-1

Table~\ref{tab:toys4k_3d_metrics} presents our quantitative evaluation. Consistent with our previous observations, our pre-trained model struggles with the domain shift, yielding lower metrics. Visually (Figure~\ref{fig:3d_asset_grid}), this manifests as severe structural artifacts, poor prompt adherence, and the hallucination of ground planes fused with the generated assets.

Domain-specific fine-tuning mitigates these issues, significantly improving quantitative metrics to approach the baseline TRELLIS text-to-3D performance. In most cases, the fine-tuned model suppresses background artifacts and captures specific details from the text prompt, such as the “yellow and black” colour scheme (Example 1), the animal shape of the robot (Example 2), or the “faded green” and “rusted roof rack” on the van (Example 4). In Appendix \ref{appendix:prompting ablation}, we compare our fine-tuning approach against explicit negative prompting, demonstrating that prompt engineering is insufficient to overcome the domain shift.

However, generation quality remains somewhat lower than the text-to-3D baseline, particularly in terms of geometric fidelity. Qualitatively, fine-grained attributes are often still missed, mirroring our image-retrieval findings. For example, the robot in Example 3 lacks the “orange and teal” colours and “blocky” shape, while artifacts such as malformed wheels persist in some cases.

\begin{figure}[t]
\centering
\setlength{\tabcolsep}{2pt}
\renewcommand{\arraystretch}{1}
\resizebox{0.8\linewidth}{!}{
\begin{tabular}{ccc}
\toprule
\textbf{Input image} & \textbf{3D Gaussian} & \textbf{Mesh} \\
\midrule

\includegraphics[width=0.33\linewidth]{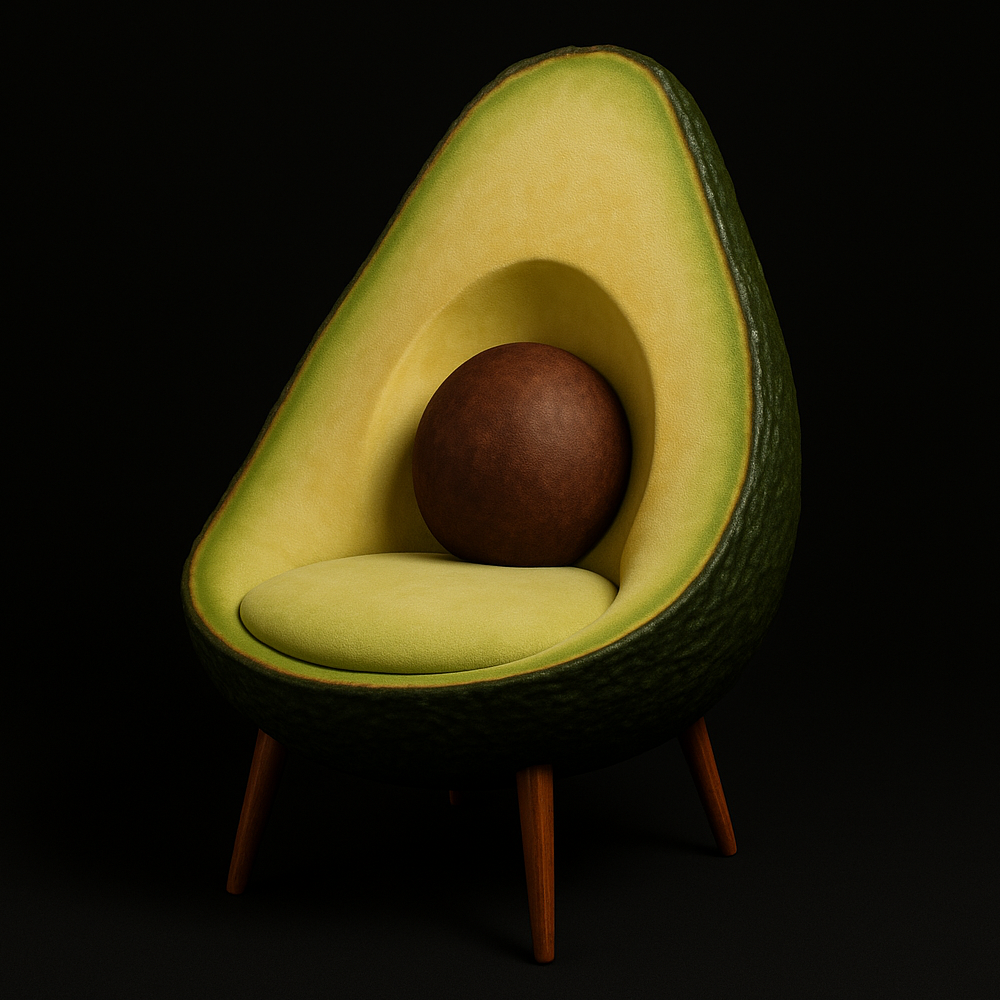} &
\includegraphics[width=0.33\linewidth]{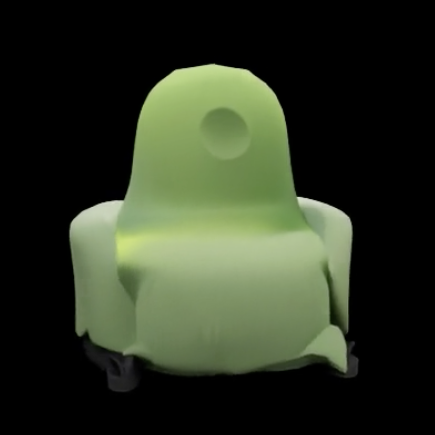} &
\includegraphics[width=0.33\linewidth]{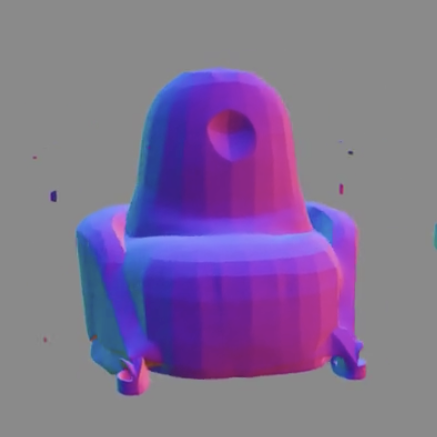} \\
\multicolumn{3}{p{1\linewidth}}{%
\footnotesize
\emph{Input caption:} ``A 3d model of this object.''
}\\[-1pt]

\includegraphics[width=0.33\linewidth]{images_2025-12-29/avocado_chair_cropped.png} &
\includegraphics[width=0.33\linewidth]{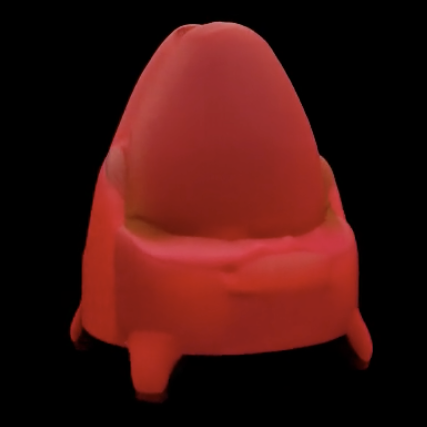} &
\includegraphics[width=0.33\linewidth]{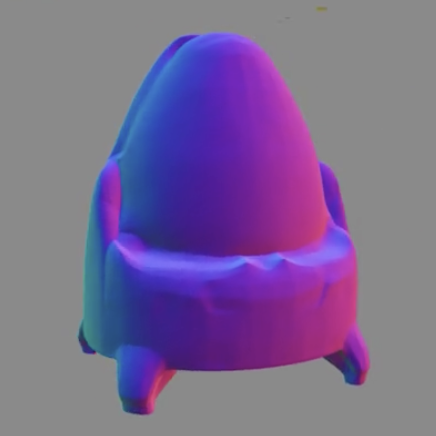} \\
\multicolumn{3}{p{1\linewidth}}{%
\footnotesize
\emph{Input caption:} ``Make it red.''
}\\[-1pt]

\includegraphics[width=0.33\linewidth]{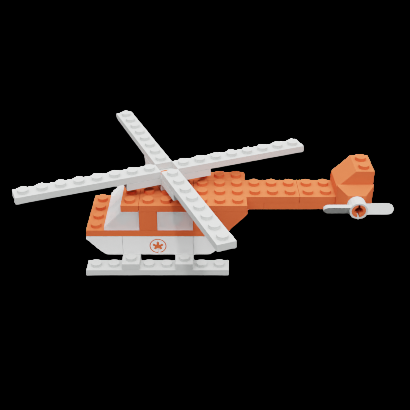} &
\includegraphics[width=0.33\linewidth]{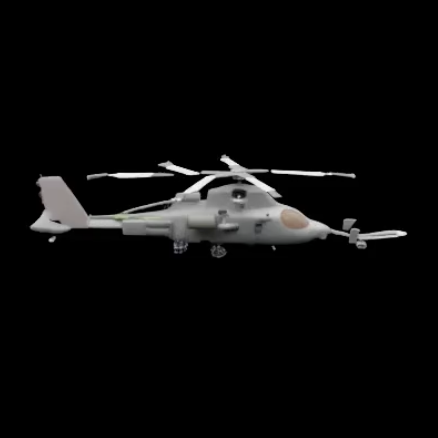} &
\includegraphics[width=0.33\linewidth]{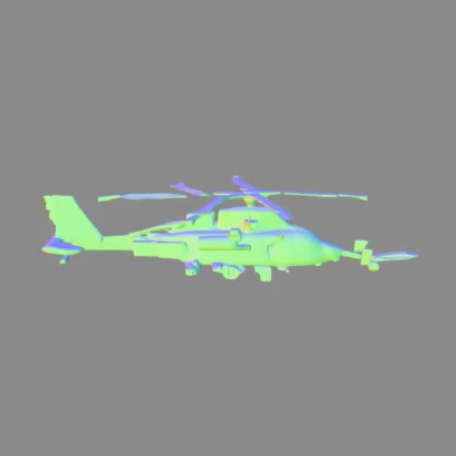} \\
\multicolumn{3}{p{1\linewidth}}{%
\footnotesize
\emph{Input caption:} ``Change from lego to normal metal.''
}\\

\end{tabular}
}

\caption{3D asset editing with fine-tuned GAP3D, using multimodal input.}
\label{fig:3col_4row_grid}
\end{figure}

\subsection{Emergent Multimodal Capabilities} \label{subsec:Emergent Multimodal Capabilities}
Although our alignment module was trained exclusively with text input, we explore whether the learned soft tokens can generalize to multimodal inputs by leveraging the joint representation space of the frozen VLM. To test this, we include both an image and a text instruction in the input prompt. Qualitative results are presented in Figure~\ref{fig:3col_4row_grid}.

The results demonstrate emergent zero-shot capabilities for multimodal generation. However, as observed earlier, the model still prioritizes high-level semantics over low-level structural detail, functioning primarily as a semantic translator rather than a structure-preserving editor. For example, when provided with an image of an avocado-shaped chair or a Lego helicopter together with textual instructions (e.g., ``\textit{Make it red}'' or ``\textit{Change from lego to normal metal}''), the model successfully extracts the object class and applies the requested material or colour changes. However, it does not strictly preserve the input geometry and instead generates a new asset that is aligned with the semantic concepts from both modalities, effectively treating the image as a rich semantic prompt. Although we did not attempt this due to computational resource restrictions, we believe that explicit training on image reconstruction or editing tasks could likely improve the preservation of input image geometry.\looseness-1

\section{Discussion and Conclusion}

In this work, we explored the alignment of VLM representations to the dense, high-dimensional patch-level embedding space of a pre-trained image encoder (DINOv2) via rectified flow. Evaluated on 3D asset generation using the frozen TRELLIS backbone, our modular generative alignment approach, GAP3D, achieves semantic and visual quality approaching the native TRELLIS text-to-3D baseline, without re-training the 3D generative model. To achieve this, domain adaptation proved critical for mitigating background artifacts and aligning feature distributions. Furthermore, despite being trained exclusively with text input, our model exhibits emergent zero-shot capabilities for multimodal prompting.

Despite this, challenges remain in capturing fine-grained geometry and low-level visual details. Our multimodal module currently acts as a semantic translator, often missing fine-grained detail and demonstrating reduced
geometric, and sometimes visual fidelity in comparison
to the native text-to-3D baseline. We hypothesize that this stems from the limited capacity of the learnable VLM soft tokens and a lack of pixel-level supervision during pre-training. Additionally, while domain-specific fine-tuning resolves background artifacts and improves overall quality, it induces catastrophic forgetting of general-domain data.  Future work could explore increasing the number of image latents, incorporating reconstruction-based pre-training objectives, and mixing datasets during fine-tuning, to increase representation capacity, provide stronger signals for high-frequency spatial details, and improve generalization.

Ultimately, our findings establish the feasibility of mapping VLM-encoded latents into the dense representation space of a pre-trained image encoder via diffusion-based generative alignment. We hope this motivates further research into modular foundation model integration, where encoders and generators can be switched out or upgraded independently to achieve scalable, versatile, and high-quality generation.

\section*{Impact Statement}
This paper presents work whose goal is to advance the field of Machine
Learning. There are many potential societal consequences of our work, none of
which we feel must be specifically highlighted here.

\section*{Acknowledgements}
We thank Professor Cees G. M. Snoek for providing access to computational resources and for his support during this work.

\bibliography{example_paper}
\bibliographystyle{icml2026}

\newpage
\appendix
\onecolumn

\section{Extended Dataset and Implementation Details}

\subsection{Dataset Preparation and Augmentation} \label{appendix:data-generation}

\subsubsection{Pre-training Data} 
For the initial general-domain pre-training phase, we utilize the public BLIP3-o dataset, which comprises approximately 25 million images paired with detailed synthetic captions (averaging 120 tokens, generated by Qwen2.5-VL-7B-Instruct) and an additional 6 million images paired with concise captions (averaging 20 tokens).

\subsubsection{Domain-Specific Fine-tuning Data}
To generate 3D renderings for fine-tuning our model, we utilize the Objaverse-XL public dataset \cite{deitke2023objaversexluniverse10m3d}. We randomly sample 60k 3D assets from this dataset and select one view per asset, chosen to minimize occlusion. For this purpose, we uniformly sample 5000 points from the 3D mesh of each asset. We then evaluate 4 candidate camera positions at uniform angles of $0^\circ$, $90^\circ$, $180^\circ$, and $270^\circ$. For each candidate, we cast rays from the camera to the sampled points, counting the number of points visible to the camera. We select the angle with the maximum visibility and render the final image from this viewpoint. We note that we utilize the open-source scripts provided by TRELLIS to download and render the 3D assets, adapting the code to support minimum occlusion view selection. 

To pair the resulting rendering with a caption, during training we perform random sampling among all available captions per asset, which serves as a form of data augmentation.

\subsection{Training Hyperparameters and Setup}
\label{appendix:hyperparameters}

Our trainable generative alignment module is a Diffusion Transformer (DiT) based on the Lumina-Next architecture \cite{zhuo2024luminanextmakingluminat2xstronger}, consisting of 36 transformer blocks with a hidden dimension of $D_{DiT}=1792$. This results in approximately
4.75 billion parameters for our VLM and diffusion module,
out of which approximately 1.36 billion are trainable.

We train our model using the AdamW optimizer \cite{loshchilov2018decoupled} 
with a per-GPU batch size of 16, utilizing 128 AMD MI250 GPUs.

The training pipeline consists of two stages:
\begin{itemize}
    \item \textbf{Stage 1 (Pre-training):} The model is trained on the 
    general-domain BLIP3-o image-text pairs for 3 epochs, using a cosine 
    annealing learning rate schedule that decays from $2.8 \times 10^{-4}$ 
    to $1 \times 10^{-5}$.
    \item \textbf{Stage 2 (Fine-tuning):} The model is fine-tuned exclusively 
    on the curated Objaverse-XL 3D asset renderings for 1000 epochs, using a 
    cosine annealing schedule that decays from $5.6 \times 10^{-4}$ to $0$.
\end{itemize}

During both stages, we apply the following training loss weights: $\lambda_p=0.4$ for the patch grid, $\lambda_{cls}=0.3$ for the global class token, and $\lambda_{reg}=0.3$ for the register tokens.

\section{Architecture} \label{appendix:architecture}

\begin{figure}[t]
    \centering
    
        \includegraphics[width=0.9\textwidth]{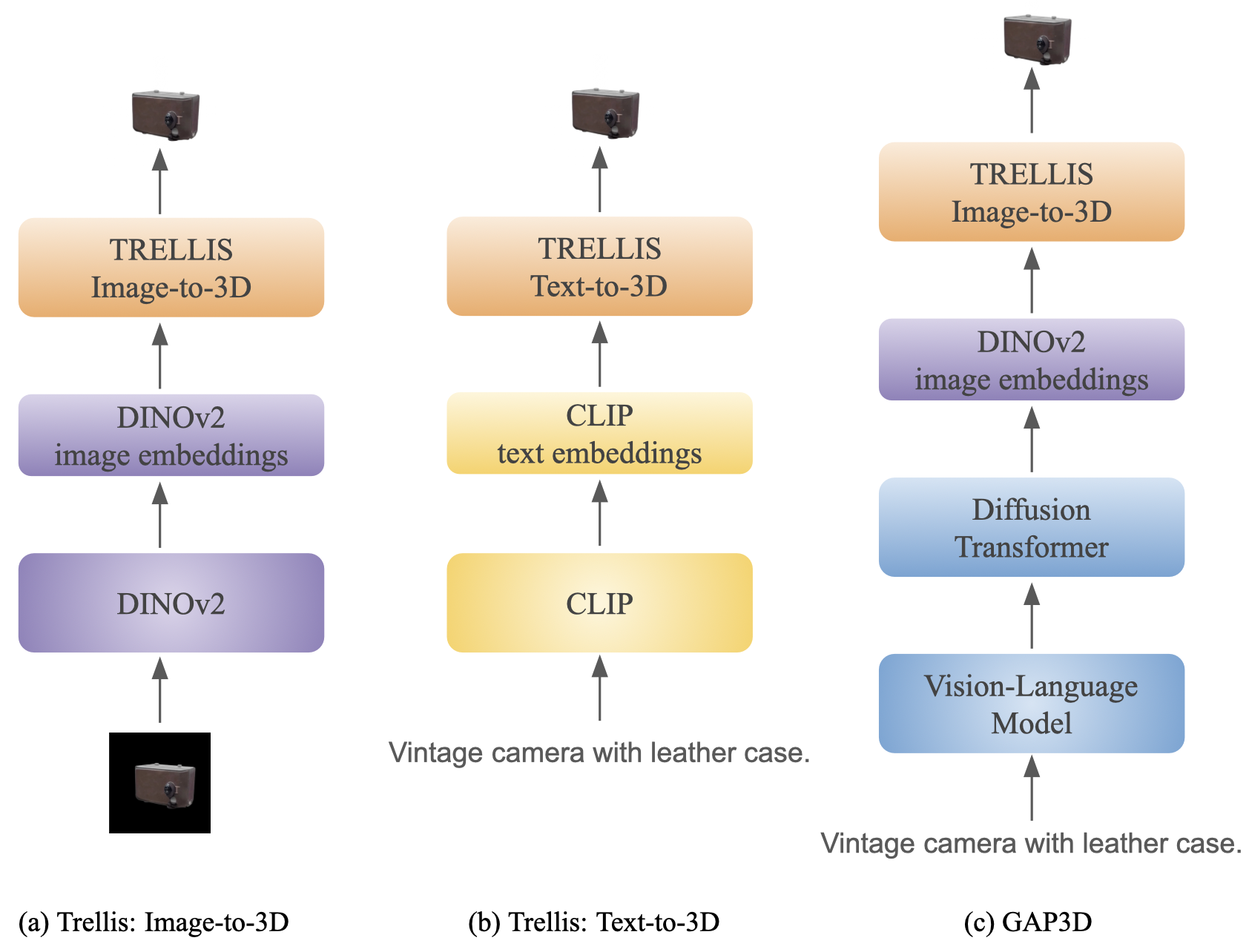}
        \label{fig:sub1}

    \caption{Overview of the TRELLIS image-to-3D (a) and text-to-3D (b) pipelines, compared to GAP3D (c). The downstream 3D generative model remains the same for our pipeline and for TRELLIS image-to-3D. However, the conditioning branch differs: We utilize a VLM followed by a diffusion alignment module that maps the VLM image latents to image embeddings. This replaces DINOv2 in the TRELLIS image-to-3D pipeline. }
    \label{fig:architecture}
\end{figure}

Figure \ref{fig:architecture}, provides a comparative overview of the conditioning pipelines for the baseline TRELLIS architectures and our proposed modular approach. There are two distinct versions of the native TRELLIS: an image-to-3D pipeline conditioned on dense, spatially structured DINOv2 features (Figure 5a), and a text-to-3D pipeline conditioned on CLIP text embeddings (Figure 5b).

As illustrated in Figure 5c, GAP3D bridges these two paradigms. We replace the DINOv2 image encoder with our VLM and generative alignment module, and map text (or multimodal) prompts directly into the dense DINOv2 feature space. This allows us to perform text-to-3D generation while leveraging the frozen image-to-3D TRELLIS backbone, aiming to bypass the limitations of compressed text embeddings.

\section{Evaluation Protocol} \label{appendix:evaluation}

To assess the alignment of the generated embeddings $\hat{\mathbf{x}}$ with the ground-truth target embeddings $\mathbf{x}$, we utilize three metrics. All metrics are computed element-wise per token and averaged across the dataset. We compute metrics individually for the CLS token and the patch tokens.

\paragraph{Cosine Similarity.}
We utilize cosine similarity to measure the directional alignment between the generated and target embeddings. For a batch of $B$ samples with $N$ tokens each (where $N$ corresponds to the patch grid size, or 1 for CLS and 4 for register tokens), the metric is defined as:
\begin{equation}
    \text{Cos}(\hat{\mathbf{x}}, \mathbf{x}) = \frac{1}{B \cdot N} \sum_{i=1}^{B} \sum_{j=1}^{N} \frac{\hat{\mathbf{x}}_{i,j} \cdot \mathbf{x}_{i,j}}{\max(\|\hat{\mathbf{x}}_{i,j}\|_2 \|\mathbf{x}_{i,j}\|_2, \epsilon)}
\end{equation}
where $\epsilon=1e^{-8}$ is a small constant for numerical stability, and $\mathbf{x}_{i,j}$ is the embedding of the j-th token in the i-th sample in the batch.

\paragraph{Mean Squared Error (MSE).}
We report the standard MSE to measure the absolute reconstruction fidelity in the feature space, penalizing discrepancies in both direction and magnitude:
\begin{equation}
    \text{MSE}(\hat{\mathbf{x}}, \mathbf{x}) = \frac{1}{B \cdot N \cdot D} \sum_{i=1}^{B} \sum_{j=1}^{N} \| \hat{\mathbf{x}}_{i,j} - \mathbf{x}_{i,j} \|_2^2
\end{equation}
where $D$ is the feature dimension (1024 for our case).

\paragraph{Norm Ratio.}
To evaluate whether the diffusion module preserves the magnitude of the feature distribution, we utilize norm ratio:
\begin{equation}
    \text{Norm Ratio} = \frac{1}{B \cdot N} \sum_{i=1}^{B} \sum_{j=1}^{N} \frac{\|\hat{\mathbf{x}}_{i,j}\|_2}{\|\mathbf{x}_{i,j}\|_2 + \epsilon}
\end{equation}
where $\epsilon=1e^{-8}$ is a small constant for numerical stability.

\subsection{3D Asset Generation Evaluation}
To evaluate the visual fidelity and semantic alignment of the generated 3D assets, we adopt the standard Fréchet Distance (FD)~\cite{heusel2018ganstrainedtimescaleupdate}, Kernel Distance (KD)~\cite{bińkowski2021demystifyingmmdgans} and CLIP Score \cite{radford2021learningtransferablevisualmodels} metrics, following the protocol established by Xiang et al.\yrcite{xiang2024structured}.

\paragraph{Data Preparation.}
We randomly sample $1,250$ assets from the Toys4K dataset. For each ground-truth asset, we render 4 canonical views using cameras positioned at yaw angles of $0^\circ, 90^\circ, 180^\circ,$ and $270^\circ$, with a fixed pitch of $30^\circ$, a field of view (FOV) of $40^\circ$, and a radius of $2.0$. This results in a reference set of $5,000$ images.

\paragraph{Generation Protocol}
For each test sample, we generate a corresponding 3D asset. For the baseline TRELLIS text-to-3D pipeline and for our proposed method we condition on the asset's caption. We randomly sample a caption when multiple captions are available. For the baseline TRELLIS image-to-3D pipeline we condition directly on a ground-truth rendering.

\paragraph{Visual Fidelity}
To assess visual fidelity, we render the generated assets using the same 4-view camera protocol as we did for the ground truth. We extract feature embeddings for both the ground-truth and generated sets using two encoders:
\begin{itemize}
    \item \textbf{InceptionV3 (Pool3):} Captures high-level semantic and perceptual similarity.
    \item \textbf{DINOv2 (ViT-L/14):} Captures structural similarity.
\end{itemize}
We report the Fréchet Distance (FD) and the unbiased Kernel Distance (KD) (scaled by $100$).

\paragraph{Geometric Fidelity }
To compute geometric quality, we generate point clouds by back-projecting depth maps from 20 views and applying Farthest Point Sampling (FPS) to obtain 4,000 points per object. We extract geometric features using a pre-trained PointNet++ classifier and report the FD between the feature distributions of the generated and ground-truth point clouds.

\paragraph{Semantic Alignment (CLIP Score)}
To evaluate semantic alignment, we render 8 views for each generated asset (with yaw intervals of $45^\circ$) and compute the average cosine similarity between the CLIP embeddings (ViT-L/14) of these 8 rendered views and the input text caption or input image for text-to-3D and image-to-3D generation respectively.

\section{Prompting vs Fine-tuning} \label{appendix:prompting ablation}

\begin{table*}[t]
\caption{Quantitative evaluation for text-to-3D asset generation using our pre-trained and fine-tuned models, as well as using our pre-trained model with explicit instructions to avoid background and other object insertion, encoded in the text prompt.}
\label{tab:prompting_ablation}
\begin{center}
\begin{small}
\begin{sc}
\begin{adjustbox}{width=\textwidth}
\begin{tabular}{l|cccccc}
\toprule
& FD (Inception) $\downarrow$
& KD (Inception) $\downarrow$
& FD (DINOv2) $\downarrow$
& KD (DINOv2) $\downarrow$
& FD (PointNet++) $\downarrow$
& CLIP $\uparrow$ \\
\midrule
Pre-trained
& 41.90 & 0.88 & 634.57 & 52.71 & 90.89 & 27.51 \\
Fine-tuned
& \textbf{24.20} & \textbf{0.16} & \textbf{329.90} & \textbf{10.65}  & \textbf{44.37} & \textbf{29.26} \\
Prompted
& 47.25 & 1.16 & 615.01 & 48.27 & 82.21 & 26.85 \\
\bottomrule
\end{tabular}
\end{adjustbox}
\end{sc}
\end{small}
\end{center}
\end{table*}

In our proposed method, we fine-tune our model on 3D asset renderings to align the generated embeddings with the target 3D distribution. Alternatively, one might attempt to bypass fine-tuning and improve 3D asset generation quality by encoding constraints such as lack of background or additional object removal directly in the text prompt (e.g., explicitly instructing the model to generate a black background and exclude objects other than the main subject). To test this, we append the phrase:  ``The background is black and no other objects are present.'' to the text prompts provided to our pre-trained model. 

As shown in Table \ref{tab:prompting_ablation}, relying on prompt engineering alone proves ineffective. Using the pre-trained model with specific exclusion instructions yields significantly worse performance than our fine-tuning approach and can in some cases even underperform compared to the baseline pre-trained model. This aligns with our earlier observations that there is a significant distribution shift when transferring from the natural image domain to 3D rendering and that our model struggles to fully bind all fine-grained visual details requested in the text prompt. Given the scarcity of single-object, black-background, uniform-lighting examples in the large-scale pre-training data, textual instructions are insufficient to override the strong scene-composition priors learned during pre-training or remove embedding noise, and may even shift the focus away from other requested detail, further reducing generation quality. 

\begin{figure*}[t]
\centering
\setlength{\tabcolsep}{2pt} 
\resizebox{1\linewidth}{!}{
\begin{tabular}{c c c c c c}
\toprule
 \textbf{Fine-tuned (G.)} & \textbf{Fine-tuned (M.)} & \textbf{Pre-trained (G.)} & \textbf{Pre-trained (M.)} & \textbf{Prompted (G.)} & \textbf{Prompted (M.)} \\
\midrule

\includegraphics[width=0.15\linewidth]{assets/robot/finetuned.png} &
\includegraphics[width=0.15\linewidth]{assets/robot/finetuned_mesh.png} &
\includegraphics[width=0.15\linewidth]{assets/robot/pretrained.png} &
\includegraphics[width=0.15\linewidth]{assets/robot/pretrained_mesh.png} &
\includegraphics[width=0.15\linewidth]{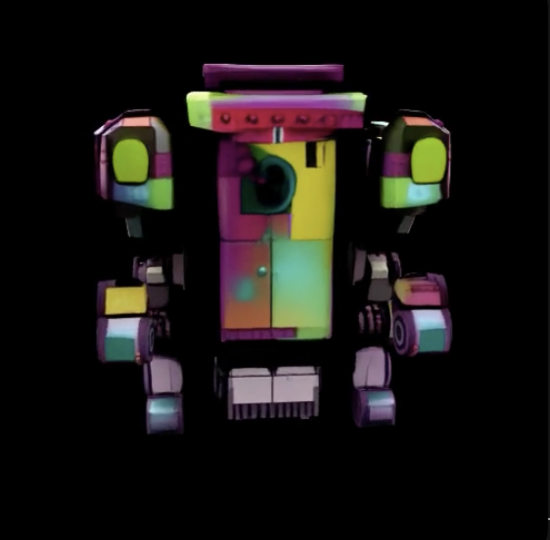} &
\includegraphics[width=0.15\linewidth]{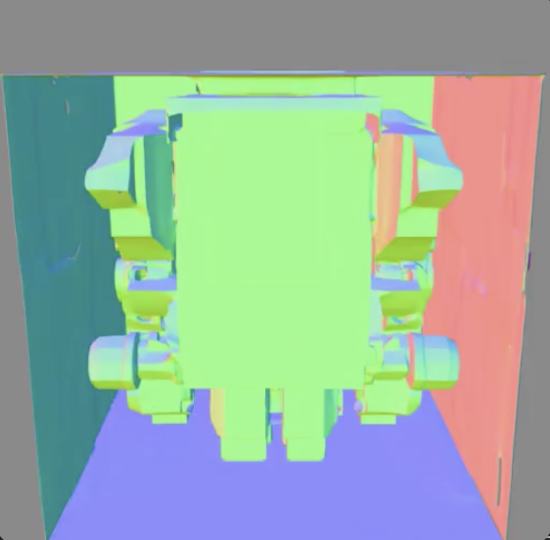} \\
\multicolumn{6}{p{0.95\linewidth}}{%
\small
\emph{Example 1:}
"Blocky, orange and teal robot with articulated limbs."
}\\
\addlinespace[4pt]

\includegraphics[width=0.15\linewidth]{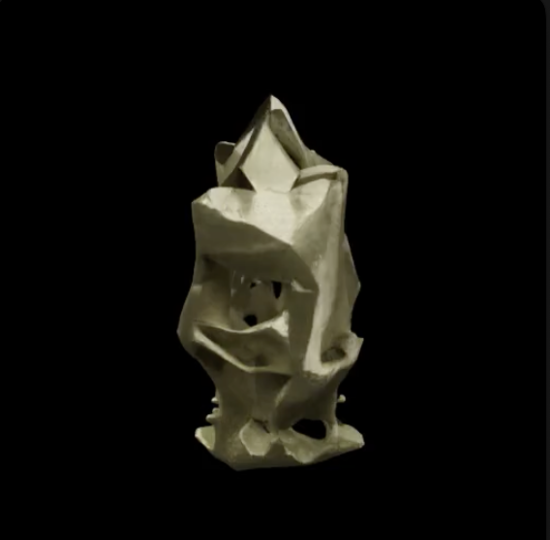} &
\includegraphics[width=0.15\linewidth]{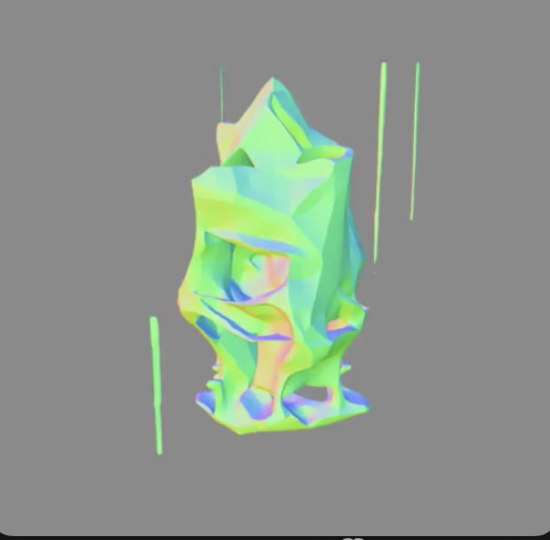} &
\includegraphics[width=0.15\linewidth]{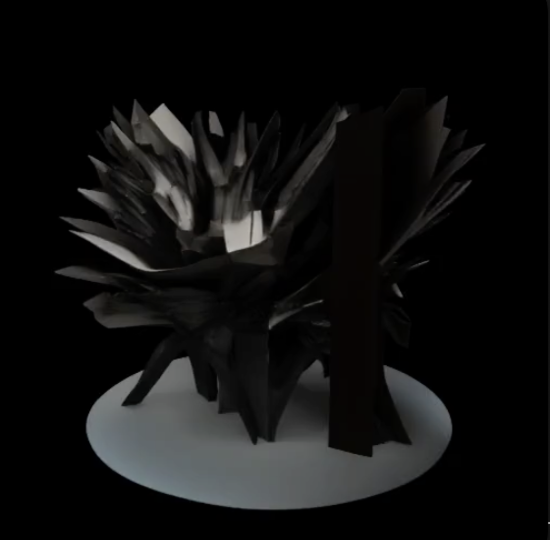} &
\includegraphics[width=0.15\linewidth]{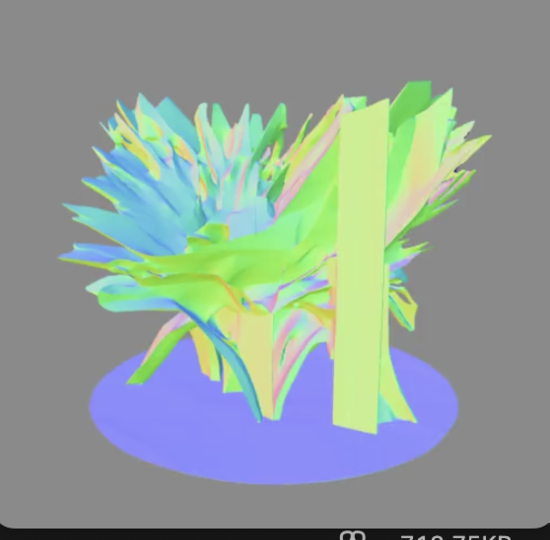} &
\includegraphics[width=0.15\linewidth]{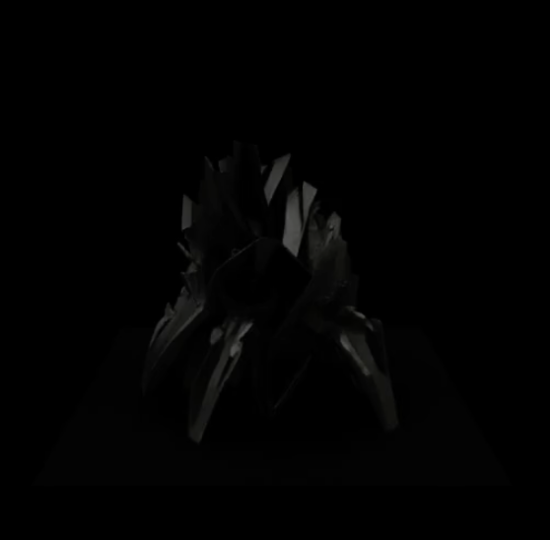} &
\includegraphics[width=0.15\linewidth]{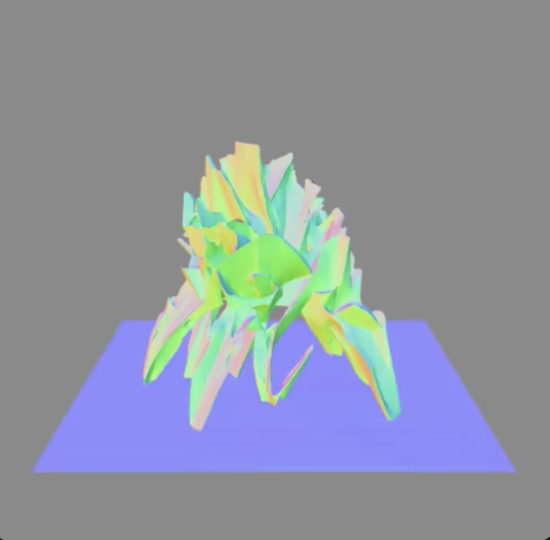} \\
\multicolumn{6}{p{0.95\linewidth}}{%
\small

\emph{Example 2:}
"Geometric metal sculpture with angular edges."
}\\

\addlinespace[4pt]

\includegraphics[width=0.15\linewidth]{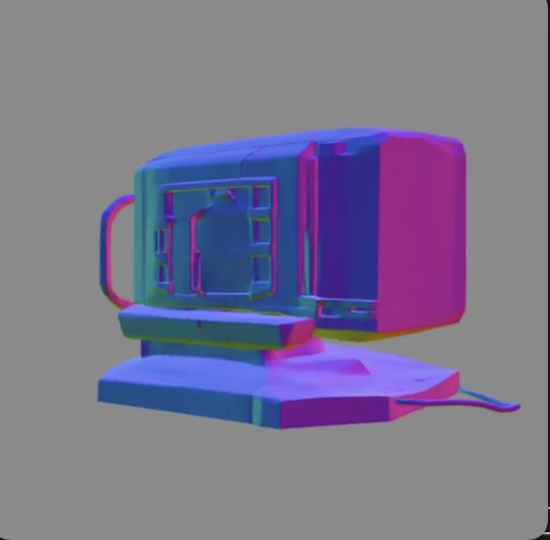} &
\includegraphics[width=0.15\linewidth]{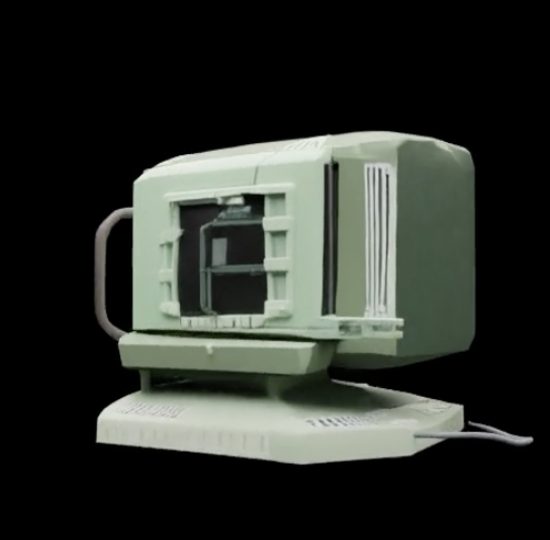} &
\includegraphics[width=0.15\linewidth]{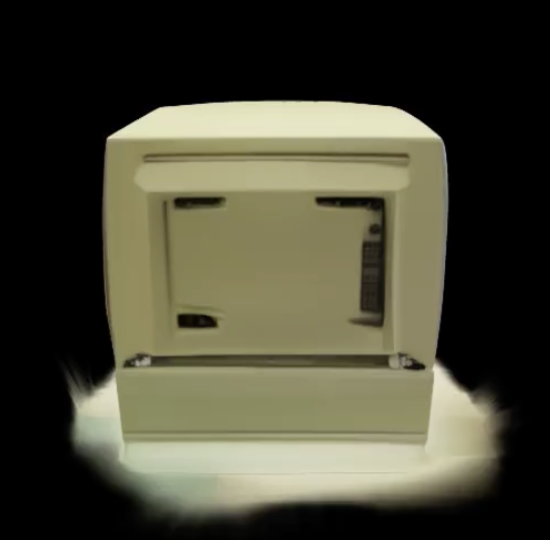} &
\includegraphics[width=0.15\linewidth]{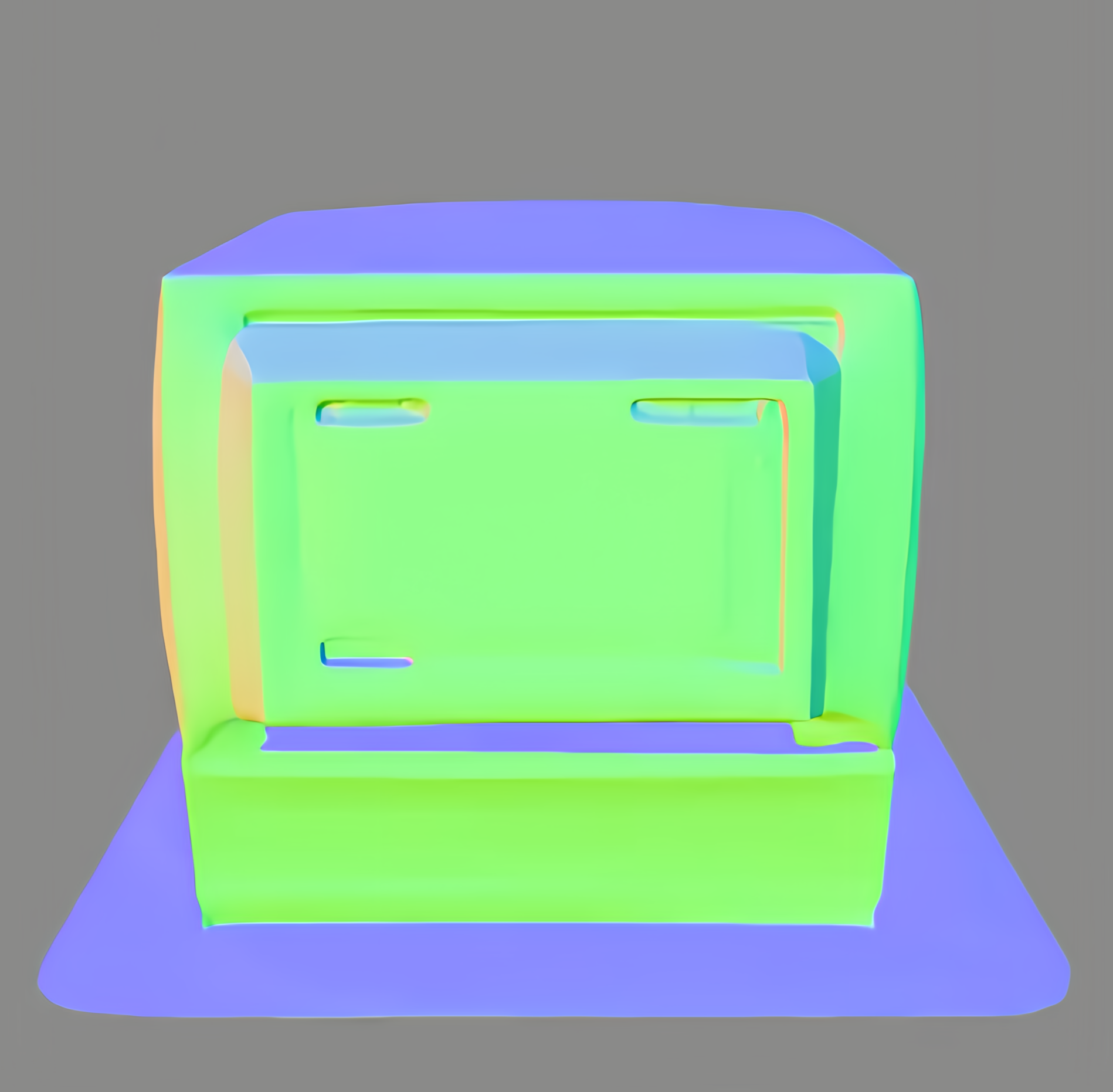} &
\includegraphics[width=0.15\linewidth]{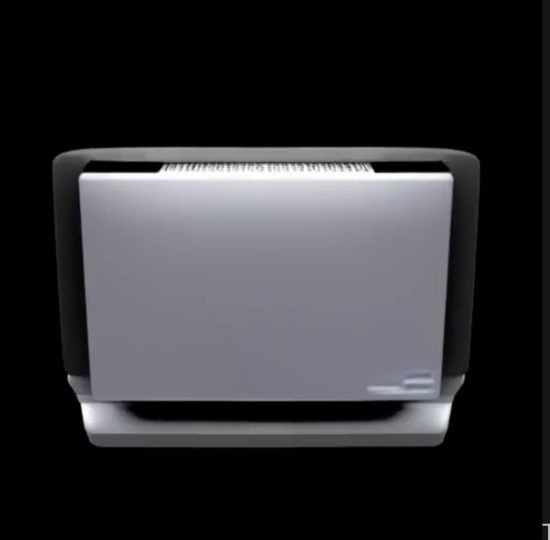} &
\includegraphics[width=0.15\linewidth]{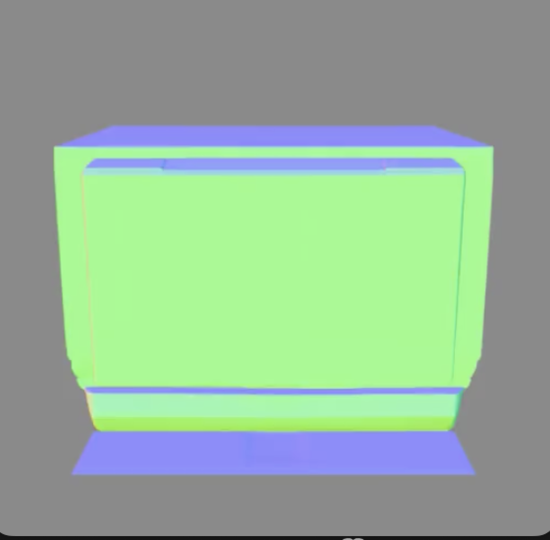} \\

\multicolumn{6}{p{0.9\linewidth}}{%
\small
\emph{Example 3:}
"Vintage green computer monitor."
}\\

\addlinespace[4pt]

\includegraphics[width=0.14\linewidth]{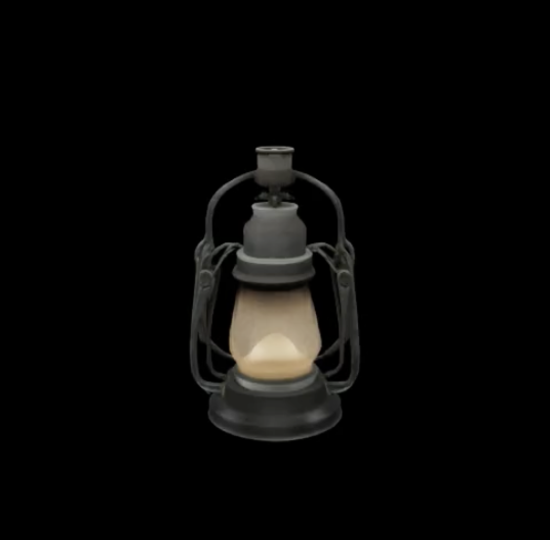} &
\includegraphics[width=0.14\linewidth]{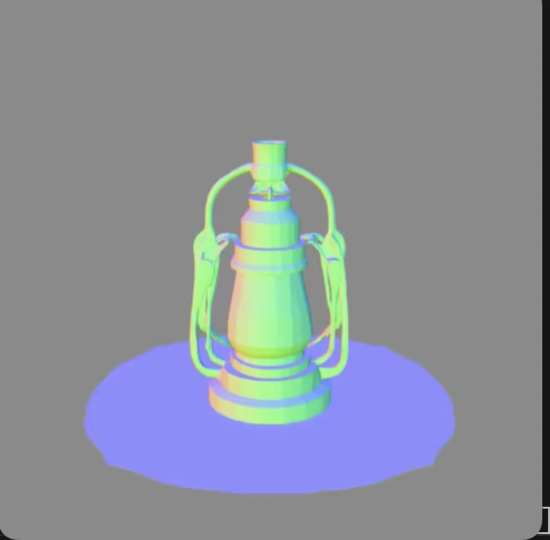} &
\includegraphics[width=0.14\linewidth]{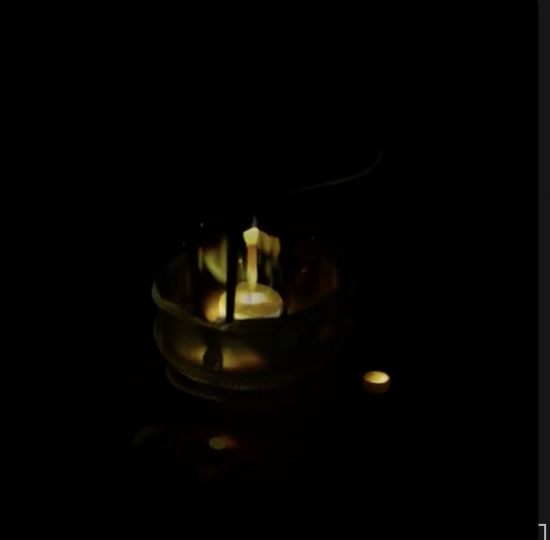} &
\includegraphics[width=0.14\linewidth]{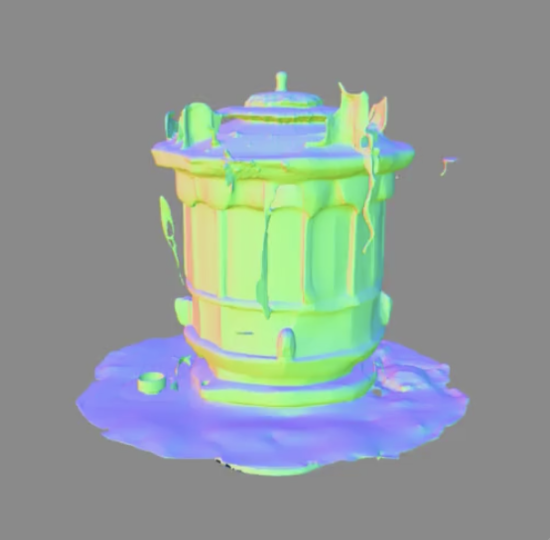} &
\includegraphics[width=0.14\linewidth]{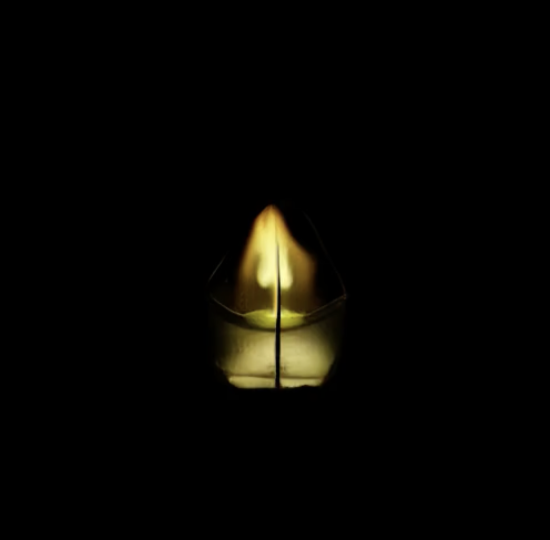} &
\includegraphics[width=0.14\linewidth]{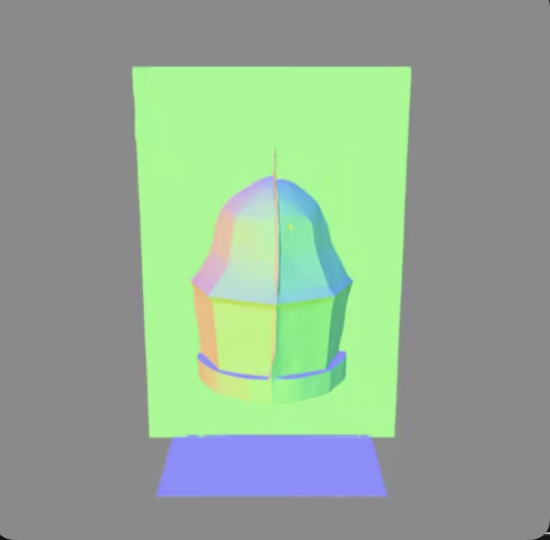} \\

\multicolumn{6}{p{0.9\linewidth}}{%
\small
\emph{Example 4:}
"Rustic lantern with a flickering flame."
}\\

\end{tabular}
}
\caption{Text-to-3D asset generation examples for our fine-tuned and pre-trained models, compared to the case of utilizing our pre-trained model while appending to the text prompt the following phrase: ``The background is black and no other objects are present.''. The 3D Gaussian (G.) and Mesh (M.) are presented for each generated asset.}
\label{fig:3d_asset_ablation_prompting}
\end{figure*}

Qualitative results (Figure \ref{fig:3d_asset_ablation_prompting}) confirm this, showing that prompt engineering fails to consistently suppress background artifacts or isolate the object. More importantly, these results indicate that the domain shift extends beyond simple environmental artifacts. For instance, in Example 4 (Lantern), the fine-tuned model still generates a floor beneath the object, yet the visual and geometric quality of the lantern itself is significantly higher compared to its pre-trained counterparts. This demonstrates that fine-tuning does not merely remove the background, but further aligns the feature distribution with the target image encoder's representation space, reducing overall noise and uncertainty.

\end{document}